\documentclass[acmtog]{acmart}
\usepackage{booktabs}

\usepackage{adjustbox}
\usepackage{arydshln}
\usepackage{cleveref}
\usepackage{geometry}
\usepackage{multirow}
\usepackage{subcaption}

\makeatletter
\def\blfootnote{\xdef\@thefnmark{}\@footnotetext}
\makeatother

\begin{document}

\title{Modulating Pretrained Diffusion Models for Multimodal Image Synthesis}

\author{Cusuh Ham}
\email{cusuh@gatech.edu}
\affiliation{%
  \institution{Georgia Institute of Technology}
  \country{USA}
}
\authornote{Work performed during an internship at Adobe Research.}

\author{James Hays}
\affiliation{%
  \institution{Georgia Institute of Technology}
  \country{USA}}
\email{hays@gatech.edu}

\author{Jingwan Lu}
\affiliation{%
  \institution{Adobe Research}
  \country{USA}}
\email{jlu@adobe.com}

\author{Krishna Kumar Singh}
\affiliation{%
  \institution{Adobe Research}
  \country{USA}}
\email{krishsin@adobe.com}

\author{Zhifei Zhang}
\affiliation{%
  \institution{Adobe Research}
  \country{USA}}
\email{zzhang@adobe.com}

\author{Tobias Hinz}
\affiliation{%
  \institution{Adobe Research}
  \country{USA}
}
\email{thinz@adobe.com}

\renewcommand{\shortauthors}{Ham, et al.}

\begin{abstract}
  We present multimodal conditioning modules (MCM) for enabling conditional image synthesis using pretrained diffusion models. Previous multimodal synthesis works rely on training networks from scratch or fine-tuning pretrained networks, both of which are computationally expensive for large, state-of-the-art diffusion models. Our method uses pretrained networks but \textit{does not require any updates to the diffusion network's parameters}. MCM is a small module trained to modulate the diffusion network's predictions during sampling using 2D modalities (e.g., semantic segmentation maps, sketches) that were unseen during the original training of the diffusion model. We show that MCM enables user control over the spatial layout of the image and leads to increased control over the image generation process. Training MCM is cheap as it does not require gradients from the original diffusion net, consists of only $\sim$1\% of the number of parameters of the base diffusion model, and is trained using only a limited number of training examples.  We evaluate our method on unconditional and text-conditional models to demonstrate the improved control over the generated images and their alignment with respect to the conditioning inputs.
  \blfootnote{\url{https://mcm-diffusion.github.io}}
\end{abstract}

\begin{teaserfigure}
  \includegraphics[width=\textwidth]{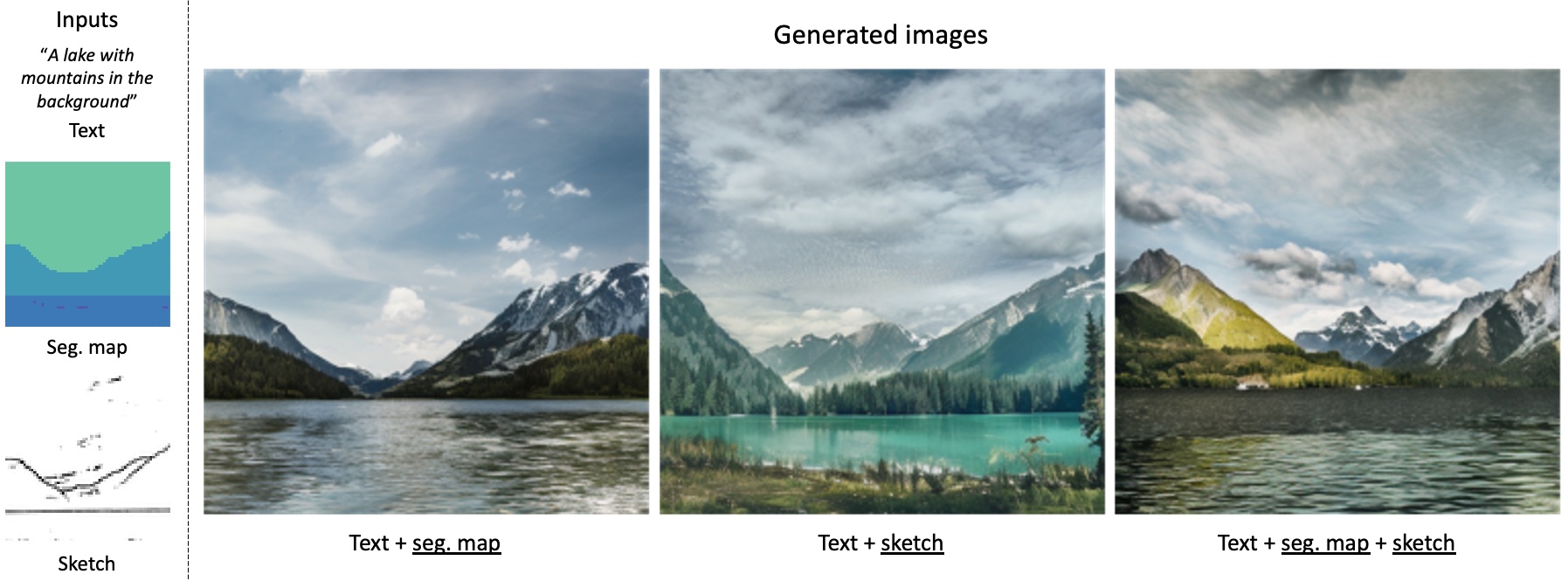}
  \vspace{-2em}
  \caption{Multimodal conditioning modules (MCMs) enable multimodal image synthesis using pretrained diffusion models. Our approach involves training a lightweight modulation network while keeping the diffusion model weights frozen. We visualize examples of adding two new modalities (underlined), segmentation maps and sketches, as conditioning modalities to Stable Diffusion, a pretrained text-conditional latent diffusion model \cite{rombach2022high}.}
  \label{fig:teaser}
\end{teaserfigure}

\maketitle


\section{Introduction}

Diffusion models have shown great potential in generating high-quality images that are realistic and diverse. However, current models rely heavily on large amounts of training data and are usually unconditional or only conditioned on more abstract conditions such as text \cite{saharia2022photorealistic, rombach2022high, ramesh2022hierarchical}. The process of training these models is expensive and requires a large amount of computational resources. The reliance on vast amounts of training data limits the models' applicability when less data is available, as is the case for many conditional generation tasks. While there exist some large datasets for text-conditional image synthesis \cite{schuhmann2022laion}, datasets for more controlled image synthesis, such as conditioning on segmentation maps, are orders of magnitudes smaller \cite{OpenImages, OpenImagesSegmentation, lin2014microsoft}.

Many approaches try to address these limitations by fine-tuning a pretrained model for a specific domain \cite{ruiz2022dreambooth,kawar2022imagic} or to accept additional conditioning modalities such as segmentation maps or sketches \cite{xie2022smartbrush}. However, this requires access to the model parameters and significant computational resources as gradients have to be calculated for the full model. Furthermore, fine-tuning a full model limits the applicability since the models are large and it can be difficult to easily share them. Thus, this approach does not scale since a new full-sized diffusion model is required for each new domain or combination of modalities. Another challenge with fine-tuning models is that they quickly overfit to the smaller subset of data that they are fine-tuned on. 

Training models conditioned on the chosen modality from scratch \cite{zhang2021m6, wu2022nuwa, gafni2022make, huang2022poegan} is limited by the available training data, reduces diversity, and diminishes the applicability of the trained model. Additionally, if the model needs to be conditioned on another modality, it needs to be retrained. A pretrained model can also be guided towards a desired direction at test time, e.g., by using gradients from a pretrained classifier or CLIP network \cite{liu2023more}. However, this approach slows down the sampling a the gradients must be calculated on the fly and optimized per sample.

Our approach addresses these limitations by introducing a novel method for multimodal conditional image synthesis using pretrained diffusion networks without changing any parameters of or requiring any gradients from the diffusion network itself. This means that the diffusion network can be treated as a black box and can even be accessed remotely, as the only data our approach needs are the predictions of the diffusion net for each sampling timestep. To achieve this, we train a small diffusion-like network conditioned on new modalities to modulate the original diffusion network's predictions at each sampling timestep so that the generated image follows the provided conditioning. The modulating network is the only model that is trained while the original diffusion network stays frozen, ensuring that the original diffusion network's high quality and diversity are preserved while also allowing for specific and tailored conditional image generation.

Our approach is computationally efficient as it requires fewer computational resources than training a diffusion net from scratch or fine-tuning an existing diffusion net. This is due to the small size of the modulating network and the lack of need to calculate gradients for the large diffusion net. Our approach generalizes well even when using only a small amount of training data. Other approaches such as fine-tuning, on the other hand, need much more training data or quickly suffer from overfitting. At test time, our approach does not slow down the sampling process since no gradients need to be calculated and the only computational overhead comes from running the small diffusion net, which is negligible compared to running the large diffusion net.

\Cref{fig:teaser} shows results of combining our multimodal conditioning module (MCM) with Stable Diffusion (SD) \cite{rombach2022high} which is originally only conditioned on text. Incorporating our MCM adds more control to the image generation by being able to condition on additional modalities such as a segmentation map or a sketch on top of the existing text condition. A single trained MCM is able to handle different input conditions, e.g., in this case only a segmentation map, only a sketch, or both. While the text specifies a rough layout of the image, the additional modalities allow for much more fine-grained control over the generation process. In the case of SD, the network predicts the noise for each sampling time step conditioned on the text and them MCM modulates the noise prediction based on the new conditions.

Our main contribution is the introduction of multimodal conditioning modules (MCM), a method for adapting pretrained diffusion models for conditional image synthesis without changing the original model's parameters. MCM is a small network trained on limited paired examples of the target modalities to modulate the output of the diffusion model during sampling. Our MCMs are roughly 100 times smaller than the original diffusion models and even when training on only a few thousand labeled examples our method obtains high-quality and diverse results while being cheaper and using less memory than training from scratch or fine-tuning a large model.


\section{Related Work}

\textbf{Conditional image synthesis.} Previous studies on conditional image synthesis have explored GANs to bridge two statistically distinct domains, such as mapping sketches or segmentation maps into photo-realistic images. One notable example is the StyleGAN series \cite{karras2019style,karras2019stylegan2,Karras2020ada,Karras2021,sauer2022stylegan}, which has served as a source of inspiration for many other conditional generation works. Another is the pix2pix series \cite{isola2017image}, including works such as \cite{park2019semantic,richardson2021encoding,sushko2022oasis}. The introduction of transformers \cite{esser2021taming} has further enhanced the visual quality of generated images.

Recently, diffusion models have emerged as an alternative to GANs and transformers, showing increased image quality and alignment with textual conditions \cite{nichol2021improved,balaji2022ediffi,feng2022ernie}.
These models have made significant advancements in text-to-image generation, with DALL-E2 \cite{ramesh2022hierarchical} proposing a framework using CLIP latent based on its previous works like GLIDE \cite{nichol2021glide} and Guided-diffusion \cite{dhariwal2021diffusion}. Latent diffusion models (LDM) \cite{rombach2022high} learn operate in the latent space of an image autoencoder, showing strong adaptability and superior quality for tasks such as segmentation-conditioned image synthesis, image super-resolution, and image inpainting. Imagen \cite{saharia2022photorealistic} uses a pyramid approach to generate high-quality images in the pixel space, marking a breakthrough in pixel-based diffusion models. There are also other pioneer works: SDEdit \cite{meng2022sdedit} proposes a stochastic differential equation during the sampling process for image editing, Diff-AE \cite{preechakul2022diffusion} conducts attribute interpolation using a diffusion model, and SR3 \cite{saharia2022image}, Palette \cite{saharia2022palette}, PITI \cite{wang2022pretraining}, and Plug-and-Play \cite{tumanyan2022plug} propose various methods for image-to-image translation using diffusion models. Additionally, ControlNet \cite{zhang2023adding}, T2I \cite{mou2023t2i}, and Latent Edge Predictor \cite{voynov2022sketch} are concurrent work that add new conditioning modalities to pretrained diffusion models.

\textbf{Multimodal conditional image synthesis.} Multimodal conditional image synthesis is a technique that uses multiple conditions from various modalities, such as masks, sketches, and language, to generate images. 
PoE-GAN \cite{huang2022poegan} uses product-of-experts GANs to synthesize images based on any subset of multiple modalities, including an empty set. 
Make-A-Scene \cite{gafni2022make} utilizes the transformer to tokenize domain-specific knowledge and adapts classifier-free guidance for the transformer use case. It accepts text and scene layouts for image synthesis. 
M6-UFC \cite{zhang2021m6} also leverages the transformer and can unify any number of multi-modal controls, where both the control signals and the synthesized image are represented as a sequence of discrete tokens.

In diffusion-based methods, eDiff-I \cite{balaji2022ediffi} utilizes multiple encoders, i.e.,  both T5 and CLIP encoders, in the diffusion model to handle text, image, and layout conditions. SpaText \cite{avrahami2022spatext} introduces spatio-textual representations to condition on text and semantic layouts.
SDG \cite{liu2023more} proposes a unified framework for semantic diffusion guidance, which allows for either language or image guidance, or both. Additionally, Composer \cite{huang2023composer} is concurrent work that conditions the diffusion net on global (e.g., text) and local (e.g., edgemaps) modalities.
The main difference between these approaches and ours is that all of them are trained from scratch on different conditioning modalities. In contrast, MCM adds new conditioning modalities to an existing model without having to retrain or fine-tune the underlying generative model itself.


\section{Approach}

We propose the \textit{multimodal conditioning module} (MCM), which aims to inject user control into pretrained diffusion models using a set of modalities originally unseen during training. MCM is a small module that is trained using limited paired examples to modulate the diffusion denoising process. We highlight several advantages of our approach: MCM 1) does not update the parameters of the diffusion model, 2) can easily be expanded to incorporate additional modalities through concatenation, 3) does not require individual modality encoders, and 4) can be applied to unconditional and conditional diffusion models. In this section, we establish notation with a brief overview of diffusion models and describe our proposed method.

\subsection{Diffusion models}
A diffusion model \cite{sohl2015deep, ho2020denoising} is trained with a defined variance schedule $\{\beta_t\}_{t=1}^T$ across $T$ timesteps. The forward noising process for an input $x_0 \in \mathbb{R}^{H \times W \times 3}$ is a fixed computation defined as:
\begin{equation}
q(x_t | x_0) = \sqrt{\overline{\alpha}_t} x_0 + \sqrt{1 - \overline{\alpha}_t} \epsilon,
\label{eq:forward-noise}
\end{equation}
where $\epsilon \sim N(0, I)$, $\alpha_t = 1 - \beta_t$, and $\overline{\alpha}_t = \prod_{i=1}^t \alpha_i$.

The reverse denoising process $p_\theta(x_t, t)$ is trained to predict the noise $\epsilon_t$ added in the forward process at timestep $t$. For diffusion models already conditioned on a given modality $y^\ast$, we use its respective encoder $\tau(\cdot)$ to encode $y^\ast$ and feed it as additional input to the reverse process, i.e., $p_\theta\big(x_t, t, \tau(y^\ast)\big)$. The reverse process is trained by optimizing the mean-squared error (MSE) between the predicted noise $\epsilon_t$ and $\epsilon \sim N(0,I)$.

Given a noisy sample $x_t$ and the predicted $\epsilon_t$, the fully denoised sample $x_0'$ can then be approximated by:
\begin{equation}
x_0' = \frac{x_t - \sqrt{1 - \overline{\alpha}_t} \epsilon_t}{\sqrt{\overline{\alpha}_t}},
\label{eq:z0}
\end{equation}
and the next denoised timestep $x_{t-1}$ can be computed using various sampling methods, such as the DDIM \cite{song2020denoising} formulation:

\begin{equation}
x_{t-1} = \sqrt{\overline{\alpha}_{t-1}} \cdot x_0' + \sqrt{1 - \overline{\alpha}_{t-1} - \sigma_t^2} \cdot \epsilon_t + \sigma_t\epsilon,
\label{eq:zt-1}
\end{equation}
where $\sigma_t = \eta\sqrt{(1 - \overline{\alpha}_{t-1}) / (1 - \overline{\alpha}_t)} \sqrt{1 - \overline{\alpha}_t / \overline{\alpha}_{t-1}}$, $\eta \in \mathbb{R}_{\geq 0}$ is a hyperparameter, and $\epsilon \sim N(0, I)$. When $\eta=0.0$, \Cref{eq:zt-1} becomes a deterministic process, which can lead to more efficient sampling using fewer timesteps.

Diffusion models can learn to either denoise $x_t$ into an RGB image directly or can work in the latent space of an autoencoder \cite{rombach2022high}. In this work, we apply our approach to LDMs as they allow for easier high-resolution image outputs and have better publicly available models. In the case of LDMs, a pretrained autoencoder $A = \{E, D\}$, consisting of an encoder $E$ and decoder $D$, is used to first encode $x$ into its latent representation $z = E(x) \in \mathbb{R}^{h \times w \times d}$. The $x_i$'s can be directly replaced with the respective $z_i$ in the above equations, and the predicted denoised latent calculated in \Cref{eq:z0} can be decoded into an image $x_0' = D(z_0')$.

\subsection{Modulating pretrained diffusion models}

\begin{figure}
  \includegraphics[width=\linewidth]{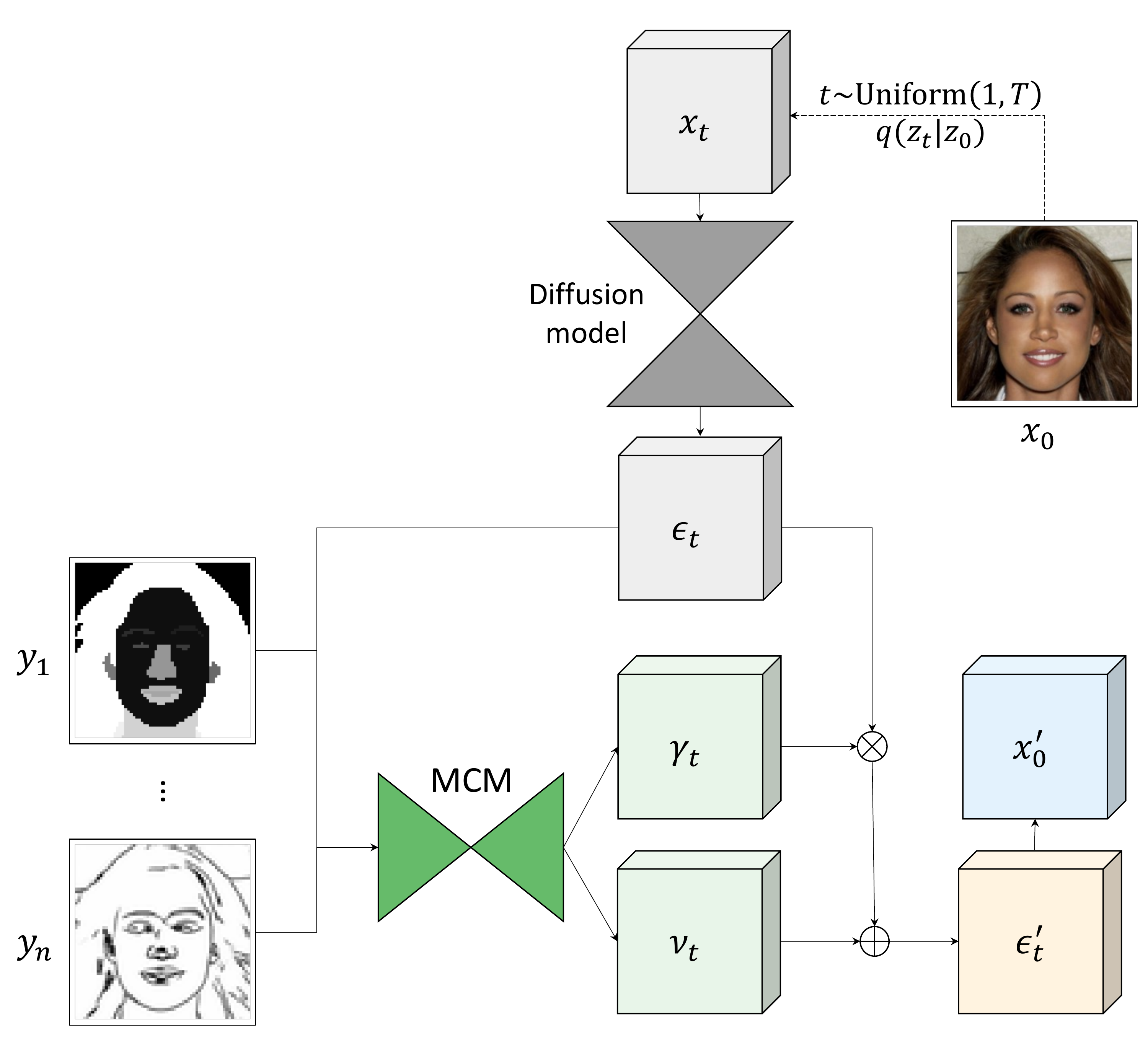}
  \caption{Illustration of our proposed modulation pipeline. Given a set of new target modalities $(y_1, ..., y_n)$, MCM predicts a set of parameters to modulate the output of a pretrained diffusion model to generate images consistent with the provided conditions.}
  \label{fig:architecture}
  \vspace{-1em}
\end{figure}

Given a paired dataset $\{(x, y_i, ..., y_n)\}$ of images $x$ and $n$ target modalities $\{y_i\}_{i=1}^n$, we train MCM, a small network that enables a pretrained diffusion model to condition its outputs on the $y_i$'s. We inject the guidance into the denoising process by using MCM to modulate the predicted noise map $\epsilon_t$ of the diffusion model at each timestep. By modulating an intermediate variable that is used to compute the next timestep $x_{t-1}$ rather than $x_{t-1}$ directly, we are not limited to using a specific sampling technique at inference.

We visualize the MCM modulation pipeline in \Cref{fig:architecture}. Similar to a standard diffusion model training step, we take an input image $x_0$. sample a random timestep $t \sim \text{Uniform}(1, T)$, compute the noised image $x_t = q(x_t | x_0)$, and get the predicted noise map $\epsilon_t = p_\theta(x_t, t)$. Given the modalities $(y_1, ..., y_n)$ corresponding to $x$, we concatenate $\{x_t, \epsilon_t, y_1, ..., y_n\}$ as input with the timestep $t$ to MCM, which outputs a set of parameters, $\{\gamma_t, \nu_t\} = \text{MCM}\big(\{x_t, \epsilon_t, y_1, ..., y_n\}, t \big)$. We use $\gamma_t$ and $\nu_t$ to modulate the predicted noise as $\epsilon_t' = \epsilon_t \otimes (1 + \gamma_t) \oplus \nu_t$. The use of spatial modulation parameters is inspired by SPADE \cite{park2019semantic}, which was originally proposed for predicting modulation parameters for normalization layers to better retain semantic information for conditional image synthesis.

We substitute $\epsilon_t$ with $\epsilon_t'$ in \Cref{eq:z0} to compute the predicted modulated denoised image $x_0'$, for which we want to adhere to the constraints specified by the $y_i$'s. The loss is defined as:
\begin{equation}
\mathcal{L}_\text{MSE} = \text{MSE}(x_0', x_0),
\label{eq:mse-loss}
\end{equation}
where $\text{MSE}(x_0', x_0)$ is the mean-squared error between the modulated denoised image $x_0'$ and the ground truth image $x_0$. For LDMs, we avoid calculating and storing gradients through the decoder $D$ by applying \Cref{eq:mse-loss} between the predicted modulated denoised latent representation $z_0'$ and ground truth latent $z_0$.

We also apply $L_1$-regularization over the modulation parameters $\gamma$ and $\nu$ to encourage $C$ to learn minimal perturbations to $\epsilon_t$:
\begin{equation}
\mathcal{L}_1 = L_1(\gamma) + L_1(\nu).
\label{eq:l1}
\end{equation}
Thus, the final training objective is defined as:
\begin{equation}
\mathcal{L}_{MCM} = \lambda_x \mathcal{L}_\text{MSE} + \lambda_1 \mathcal{L}_1,
\label{eq:loss}
\end{equation}
where $\lambda$ is a scalar weighting term.

We apply the \textit{modality dropout} technique \cite{huang2022poegan}, where, with probabilities $\{p_i\}_{i=1}^n$, we replace the respective modality $y_i$ with -1's during training. At test time, MCM is able to predict modulation parameters even in the absence of one or more modalities, avoiding heavy reliance on a single modality.

\begin{figure*}
  \centering
  \includegraphics[width=\textwidth]{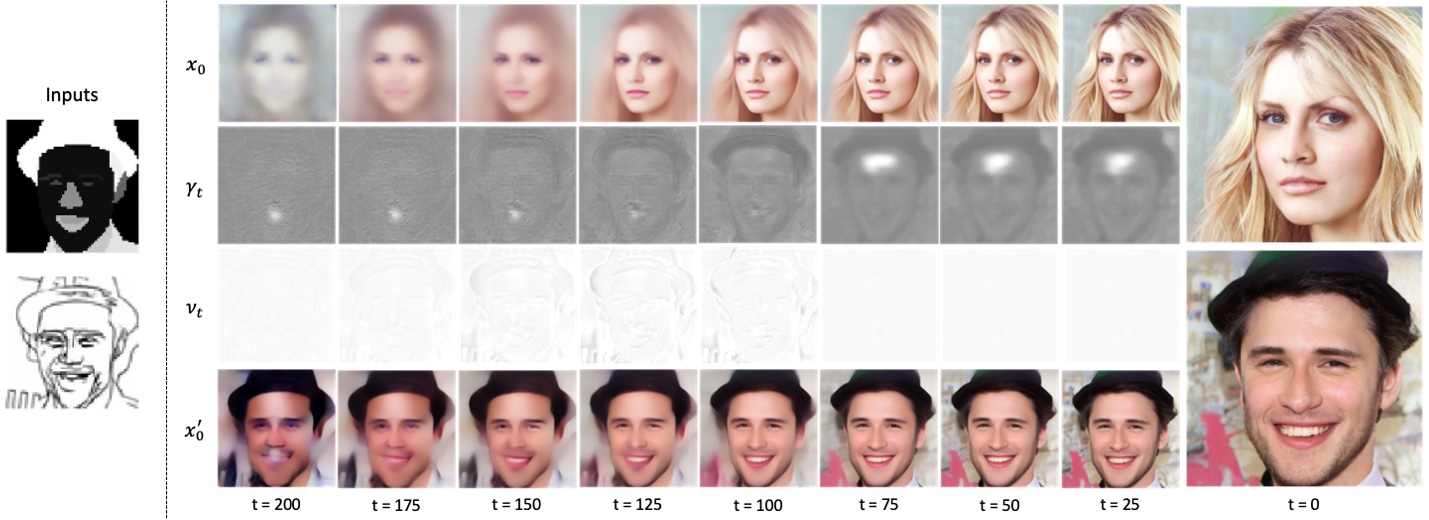}
  \caption{Given the same input noise map $z_T$, we visualize the difference in the original and modulated denoised images ($x_0$ and $x_0'$, respectively) and the modulation parameters during sampling. The magnitude of $\gamma$ and $\nu$ increases for the first half of sampling and then rapidly decreases for the remaining steps.}
  \label{fig:modulation}
\end{figure*}


\section{Experiments}
\label{sec:experiments}

In this section, we describe our experimental setup and evaluation protocols, and present qualitative and quantitative results for MCM. We primarily focus on the addition of sketches and semantic segmentation maps to latent diffusion models (LDMs) \cite{rombach2022high} due to availability of data and public model checkpoints.

\textbf{Network architectures and training details.} We leverage pretrained unconditional and text-conditioned LDMs as our base models: two unconditional LDMs trained on CelebA \cite{liu2015faceattributes} and Mountains \cite{park2020swapping} and text-conditioned Stable Diffusion v2.1 (SD) \cite{rombach2022high} trained on a subset of LAION-5B \cite{schuhmann2022laion}. The two unconditional LDMs produce $256 \times 256$ resolution images, and the text-conditioned SD produces $512 \times 512$ resolution images. We use the publicly available non-EMA weights for the CelebA and SD models, while we trained the Mountains LDM from scratch (no public checkpoints for an unconditional model were available). Experiments with MCM applied to a pixel-based diffusion model can be found in \Cref{sec:sup-pixel}.

We use a time-conditional U-Net \cite{ronneberger2015u} to output the modulation parameters, $\gamma$ and $\nu$. We train one MCM per dataset/model combination, with its number of parameters totalling $\sim$1\% of the unconditional LDMs and $\sim$0.4\% of SD. We use $\lambda_x = 1$ for unconditional LDMs and $\lambda_x = 10$ for SD. Specific architecture and training details can be found in \Cref{sec:sup-reproducibility}.

\textbf{Datasets.} We evaluate the performance of MCM one two datasets: MM-CelebA-HQ \cite{xia2021tedigan,liu2015faceattributes,karras2017progressive,CelebAMask-HQ} and Flickr Mountains \cite{park2020swapping}. MM-CelebA-HQ contains segmentation maps, sketches, and captions for 30,000 images of celebrity faces, of which $\sim$6,000 are designated test images. Flickr Mountains contains 500,000 mountain images scraped from Flickr with $\sim$6,000 test images. Because it does not contain any other corresponding modalities, we use the same pipeline used by PoE-GAN \cite{huang2022poegan} to produce pseudo-ground-truth segmentation maps and sketches: we use DeepLab-v2 \cite{chen2017deeplab} to generate segmentation maps, and HED \cite{xie2015holistically} with sketch simplification \cite{SimoSerraSIGGRAPH2016} to generate sketches. We also use BLIP \cite{li2022blip} to generate captions for Mountains for SD experiments.
Collecting paired multimodal data at the scale required to train state-of-the-art conditional generative models can be difficult and expensive. Thus, by default, we only use a randomly sampled subset of 5,000 training examples for our experiments to highlight the efficacy and application of our approach under constrained settings. We provide comparisons to MCM trained with the full CelebA dataset to quantify the effect of the amount of training data. We use the full test sets for evaluations, and visualize results generated using held-out test examples of the modalities as inputs.

\textbf{Evaluation metrics.} We use Fr\'echet Inception Distance (FID) \cite{heusel2017gans} and Learned Perceptual Image Patch Similarity (LPIPS) \cite{zhang2018unreasonable} to evaluate image quality and diversity. For each set of input modalities, we sample two images and compute the LPIPS between the two, averaged across the test set.

FID and LPIPS are qualitative metrics -- we emphasize that neither metric quantifies the alignment of the generated image to its respective conditioning inputs. However, other works on multimodal conditional synthesis only report values using qualitative metrics. We propose to use metrics from related work on conditional image editing \cite{liu2022asset} to quantify the alignment of the generated image to the conditioning inputs: 1) mean intersection over union (mIoU), 2) segmentation accuracy, and 3) sketch distance \cite{ham2022cogs}. For the two segmentation alignment metrics (mIoU and accuracy), we leverage a pretrained BiSeNet \cite{yu2018bisenet} for CelebA, and the same DeepLab-v2 \cite{chen2017deeplab} network used to generate pseudo-ground-truth segmentation maps for Mountains.

\textbf{Baselines.} Since MCM does not modify the pretrained diffusion models weights other recent GAN- or diffusion-based approaches to multimodal conditional synthesis that are trained from scratch, such as PoE-GAN \cite{huang2022poegan} and Make-A-Scene \cite{gafni2022make}, are not directly applicable as baselines. Additionally, many of these approaches do not use the same modalities explored in this work, and do not have publicly released code.

Most similar to our experimental setup is SDG \cite{liu2023more}, which leverages gradients from ``guidance'' networks for each modality to optimize each sample at test time, thus requiring a forward pass through each network at every sampling step which slows down the sampling speed. While SDG does not update the parameters of the diffusion model, the guidance networks requires fine-tuning on noisy data in order to produce meaningful gradients for the initial timesteps during sampling. Additionally, SDG was proposed for pixel-based diffusion models, but can be adapted to LDMs by performing a forward pass through the decoder at each sampling step. We omit comparisons to SDG due to memory constraints presented by the additional step through the decoder, reliance on the guidance networks, and slow sampling speeds.

Instead, we compare against fine-tuning the diffusion model directly. We expand the input channels of the first convolutional layer of the pretrained LDMs to accommodate for the additional modalities, and train using the same settings as MCM. Since we want to enable all combinations of inputs, we adjust the modality dropout rates for the fine-tuning models to $p_{seg} = p_{sketch} = p_{\{seg, sketch\}} = 0.25$. We report all metrics on unconditional samples for reference, where the unconditional outputs for MCM are directly from the original diffusion model. We use DDIM sampling \cite{song2020denoising} with $N=200$ steps and $\eta=0.0$ for all methods and evaluations, and an unconditional guidance scale of 5.0 for SD. We also include evaluations against publicly available checkpoints for segmentation- and sketch-conditioned pSp \cite{richardson2021encoding}, a StyleGAN \cite{karras2019style} encoder-based method, and to a multimodal variant of concurrent work, ControlNet \cite{zhang2023adding}.

\begin{figure*}
  \centering
  \includegraphics[width=\textwidth]{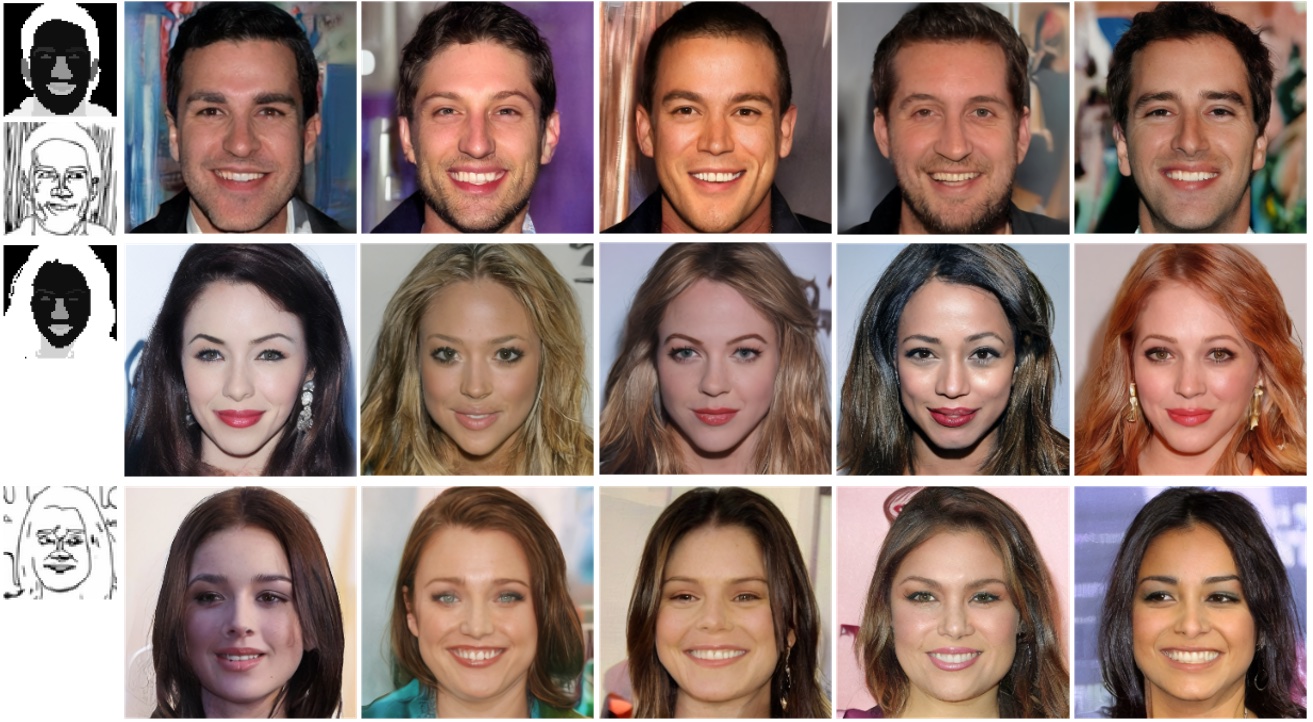}
  \vspace{-1em}
  \caption{Visualization of the diversity in MCM generation results for CelebA.}
  \label{fig:celeba-diversity}
\end{figure*}

\begin{figure*}
  \centering
  \includegraphics[width=\textwidth]{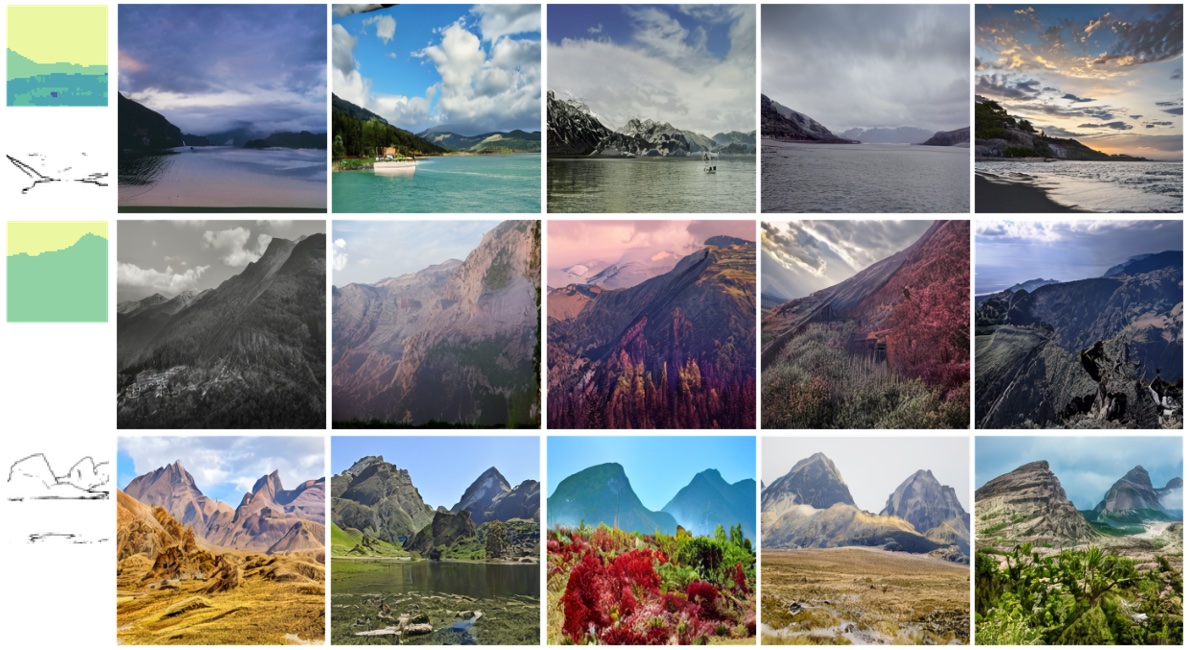}
  \caption{Visualization of the diversity in MCM generation results for Mountains.}
  \label{fig:mountains-diversity}
\end{figure*}

\begin{figure*}
  \centering
  \includegraphics[width=\textwidth]{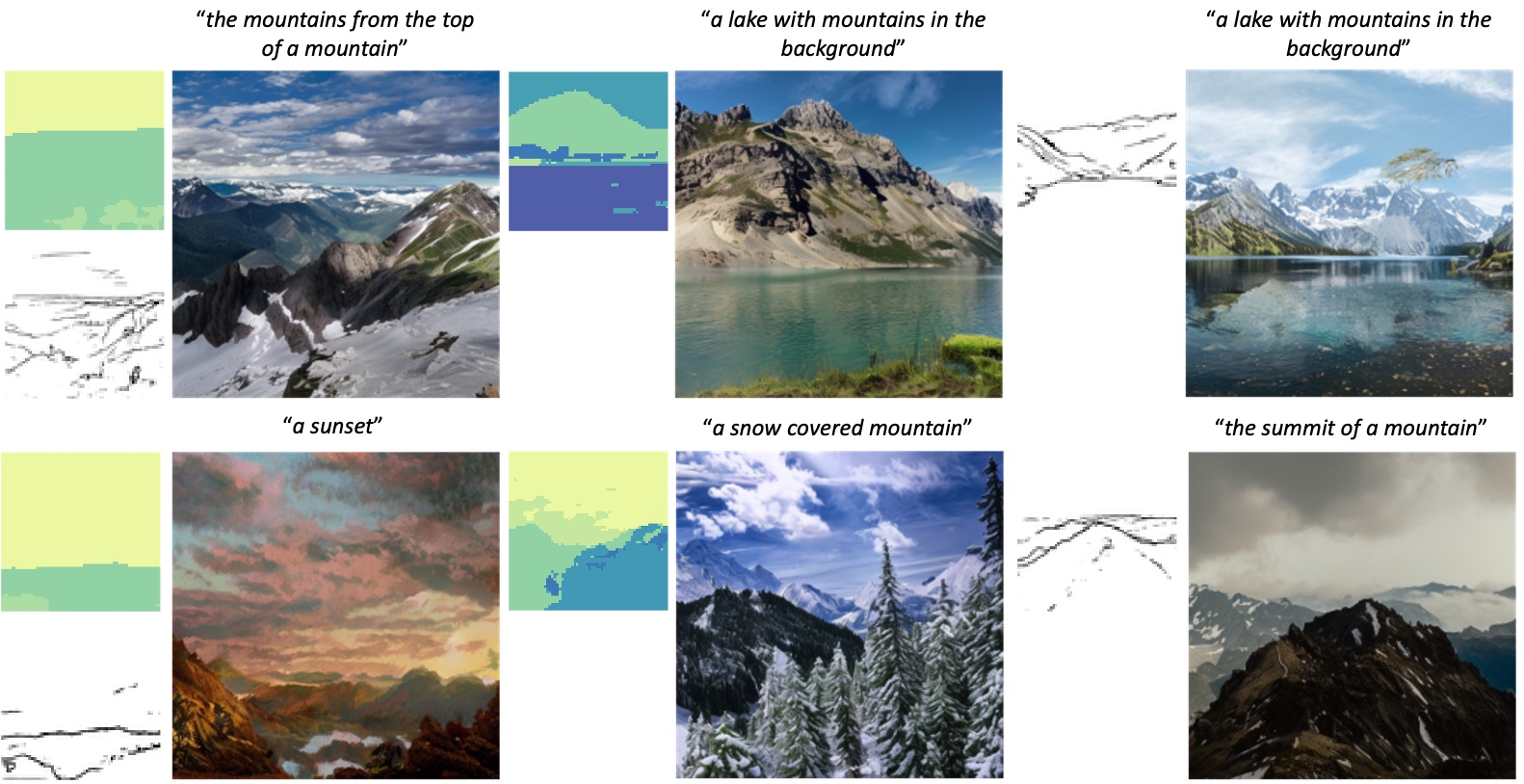}
  \vspace{-2em}
  \caption{Examples generated by MCM with Stable Diffusion for Mountains. The text prompts are directly fed as input to SD, while segmentation maps and/or sketches are inputs to MCM.}
  \vspace{-1em}
  \label{fig:mountains-sd}
\end{figure*}

\begin{table}
  \setlength{\tabcolsep}{3pt}
  \begin{adjustbox}{width=\linewidth}
    \begin{tabular}{cccccccccccc}
      \toprule
      \multirow{2}{*}{Dataset}  &  \multirow{2}{*}{Method}  &&  \multicolumn{4}{c}{FID $\downarrow$}  &&  \multicolumn{4}{c}{LPIPS $\uparrow$} \\
      &  &&  $\emptyset$  &  Seg  &  Sketch  &  Seg + Sk  &&  $\emptyset$  &  Seg  &  Sketch  &  Seg + Sk \\
      \midrule
      \multirow{4}{*}{CelebA}  &  pSp-Seg  && -  &  92.744  &  -  &  -  &&  -  &  0.341  &  -  &  - \\
      &  pSp-Sketch  &&  -  &  -  &  51.266  &  -  &&  -  &  -  &  0.326  &  - \\
      &  Fine-tune  &&  64.066  &  47.204  &  46.485  &  44.247  &&  0.474  &  0.358  &  0.288  &  0.272 \\
      &  ControlNet  &&  16.344  &  19.317  &  17.638  &  18.875 &&  0.563 &  0.496  &  0.546  &  0.492 \\ 
      &  MCM (Ours)  &&  16.344  &  18.085  &  21.065  &  18.842  &&  0.563  &  0.458  &  0.479  &  0.461 \\
      \midrule
      \multirow{2}{*}{Mountains}  &  Fine-tune  &&  102.803  &  160.096  &  104.637  &  102.341  &&  0.637  &  0.260  &  0.202  &  0.191 \\
      &  MCM (Ours)  &&  33.013  &  25.498  &  27.751  &  26.145  &&  0.752  &  0.674  &  0.664  &  0.666 \\
      \bottomrule
    \end{tabular}
  \end{adjustbox}
  \caption{Comparison of FID (quality) and LPIPS (diversity) scores.}
  \vspace{-2em}
  \label{tab:qualitative}
\end{table}

\begin{table}
  \setlength{\tabcolsep}{3pt}
  \begin{adjustbox}{width=\linewidth}
    \begin{tabular}{ccccccccccccc}
      \toprule
      \multirow{2}{*}{Dataset}  &  \multirow{2}{*}{Method}  &&  \multicolumn{3}{c}{mIoU $\uparrow$}  &&  \multicolumn{3}{c}{Accuracy $\uparrow$} & \multicolumn{3}{c}{Sketch dist. $\downarrow$}\\
      &  &  $\emptyset$  &  Seg   &  Seg + Sk  &&  $\emptyset$  &  Seg  &  Seg + Sk && $\emptyset$  &  Sketch   &  Seg + Sk  \\
      \midrule
      \multirow{3}{*}{CelebA}  &  pSp-Seg  &  -  &  0.554  &  -  &&  -  & 0.689  &  - &&  -  &  -  &  -  \\
      &  pSp-Sketch  &  -  &  -  & - &&  -  &  -  & - && - &  4.934 &  -  \\
      &  Fine-tune  &  0.405  &  0.800  &  0.824  && 0.553  &  0.877  &  0.894 &&  24.276  &  15.404  &  15.063  \\
      &  ControlNet  &  0.413  &  0.616  & 0.623  &&  0.554  &  0.723  &  0.728 && 6.225  &  7.592  &  6.072  \\
      &  MCM (Ours)  &  0.413  &  0.631  &  0.627  &&  0.554  &  0.744  &  0.741 &&  6.225  &  4.323  & 4.485  \\
      \midrule
      \multirow{2}{*}{Mountains}  &  Fine-tune  &  0.127  &  0.285  &  0.331  &&  0.181  &  0.355  &  0.399 && 41.329 & 34.323 & 34.796 \\
    &  MCM (Ours)  &  0.127  &  0.204  &  0.212  &&  0.177  &  0.359  &  0.268 && 14.177 & 7.932 & 8.527 \\
    \bottomrule
  \end{tabular}
  \end{adjustbox}
  \caption{Comparison of segmentation alignment scores.}
  \vspace{-2em}
  \label{tab:seg-sketch-alignment}
\end{table}

\textbf{Results.} We visualize the magnitude and effects of the modulation parameters on the predicted denoised images $x_0'$ during sampling using the same input noise map $z_T$ in \Cref{fig:modulation}. The unconditional predictions $x_0$ are ambiguous at larger values of $t$, whereas MCM outputs parameters that enforce more structure into $x_0'$ early on so that the final image will adhere to the inputs. The magnitude of the modulation parameters is the greater at larger values of $t$, peaking towards $t = \frac{N}{2}$, and then quickly decreasing since the remaining steps are mostly responsible for adding high frequency details to the image \cite{choi2022perception}.

\begin{figure*}
  \begin{subfigure}{.45\linewidth}
    \centering
    \includegraphics[width=\columnwidth]{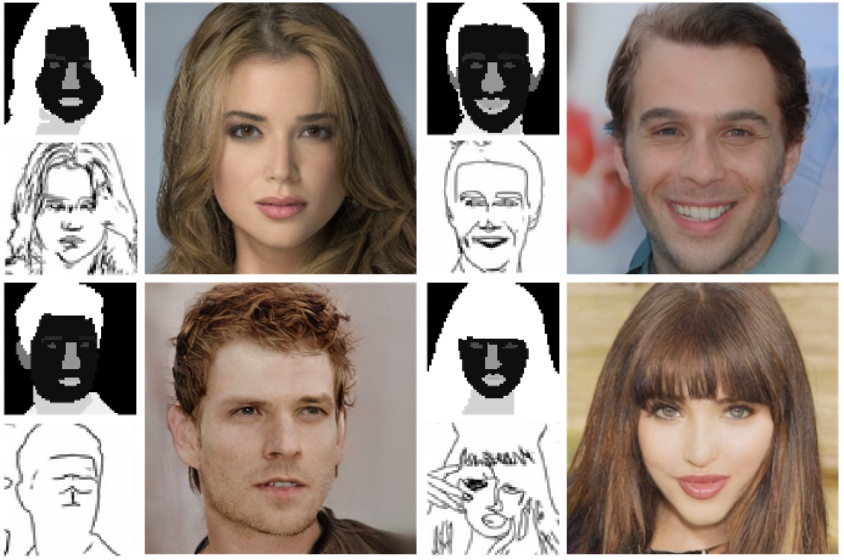}
    \caption{CelebA segmentation maps + sketches.}
    \label{fig:celeba-segmap-sketch}
  \end{subfigure}
  \begin{subfigure}{.45\linewidth}
    \centering
    \includegraphics[width=\columnwidth]{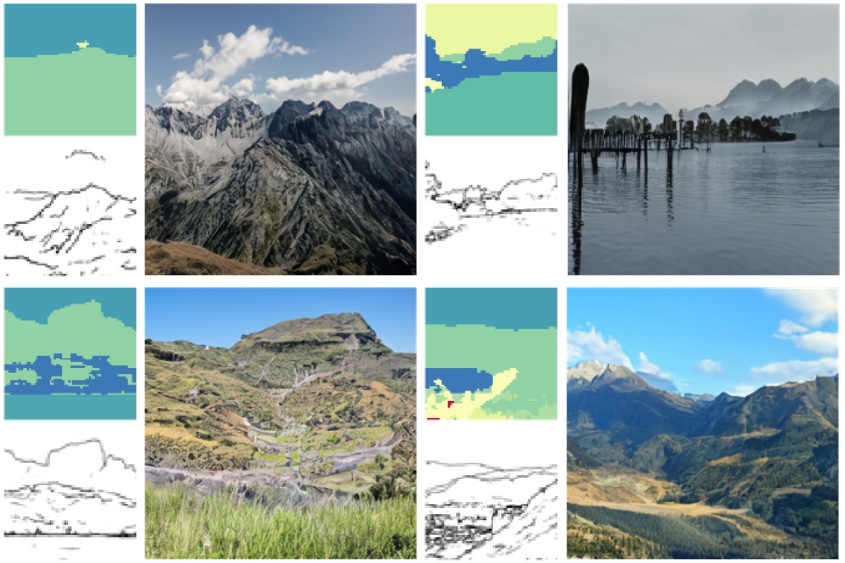}
    \caption{Mountains segmentation maps + sketches.}
    \label{fig:mountains-segmap-sketch}
  \end{subfigure}
  
  \bigskip
  \bigskip
  
  \begin{subfigure}{.45\linewidth}
    \centering
    \includegraphics[width=\columnwidth]{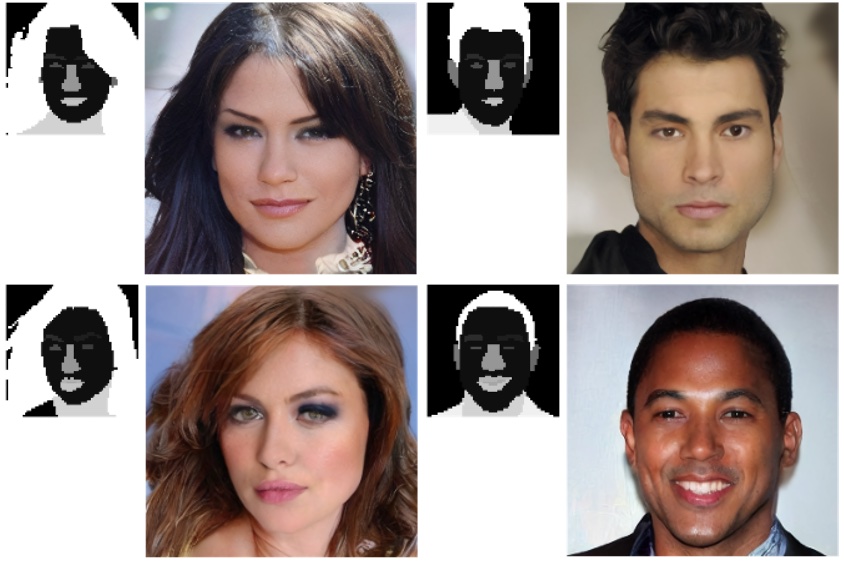}
    \caption{CelebA segmentation maps.}
    \label{fig:celeba-segmap}
  \end{subfigure}
  \begin{subfigure}{.45\linewidth}
    \centering
    \includegraphics[width=\columnwidth]{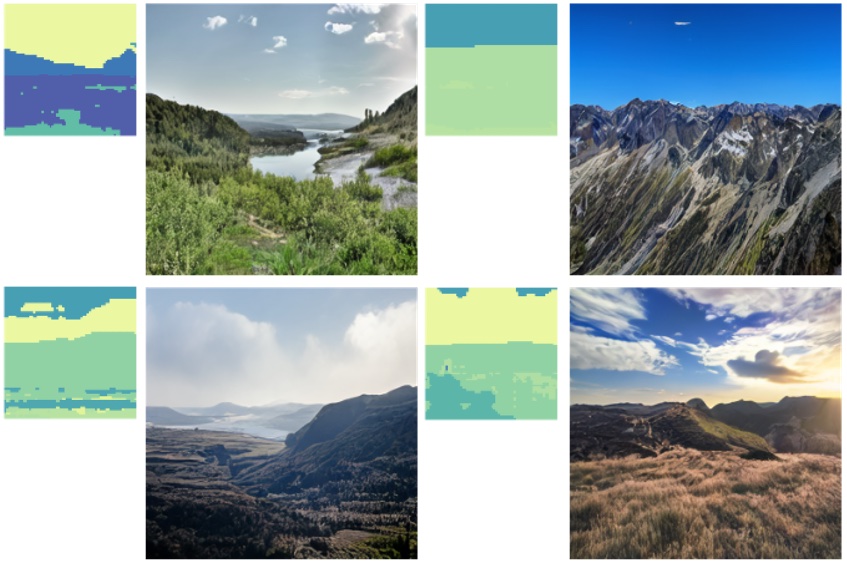}
    \caption{Mountains segmentation maps.}
    \label{fig:mountains-segmap}
  \end{subfigure}
  
  \bigskip
  \bigskip
  
  \begin{subfigure}{.45\linewidth}
    \includegraphics[width=\columnwidth]{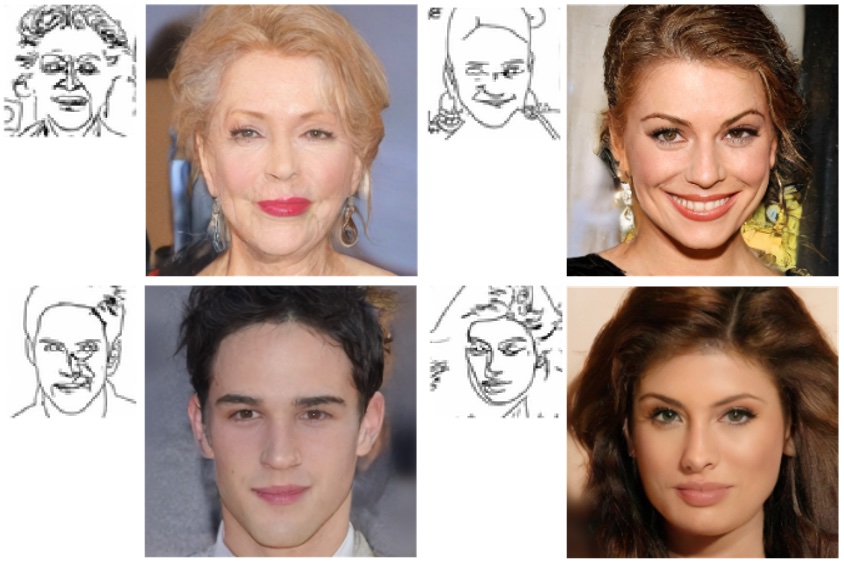}
    \caption{CelebA sketches.}
    \label{fig:celeba-sketch}
  \end{subfigure}
  \begin{subfigure}{.45\linewidth}
    \includegraphics[width=\columnwidth]{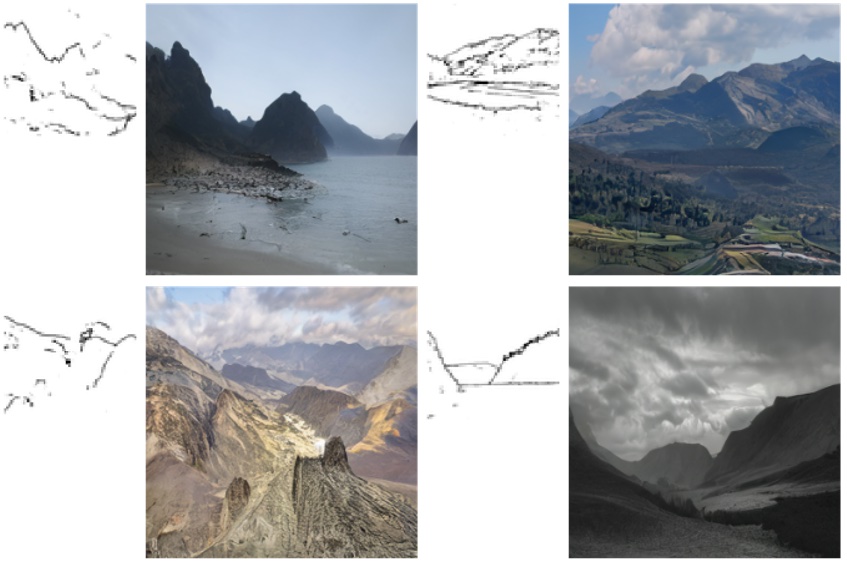}
    \caption{Mountains sketches.}
    \label{fig:mountains-sketch}
  \end{subfigure}
  \caption{Images synthesized by MCM using various combinations of inputs for CelebA and Mountains.}
  \label{fig:samples}
\end{figure*}

We show that MCM provides better control overall than the fine-tuning baseline in terms of balancing control with quality and diversity using only a small number of training examples. In \Cref{tab:qualitative}, we observe that MCM has a relatively small drop in quality and diversity from the base LDM compared to fine-tuning, and even improves the quality of the generated images for Mountains. While we expect a drop in diversity to accommodate for the constraints defined by the conditioning signals, we show that MCM is able to generate consistent yet distinct images (see \Cref{fig:celeba-diversity,fig:mountains-diversity}). Meanwhile, fine-tuning is susceptible to overfitting to a small training set, producing blurrier and less diverse images for all input combinations. We visualize more MCM examples in \Cref{fig:samples}. ControlNet performs comparably to MCM but trains a much larger model ($\sim$50\% of the size of the original diffusion model) and needs access to the original models' parameters.

\begin{figure*}
  \centering
  \includegraphics[width=0.61\textwidth]{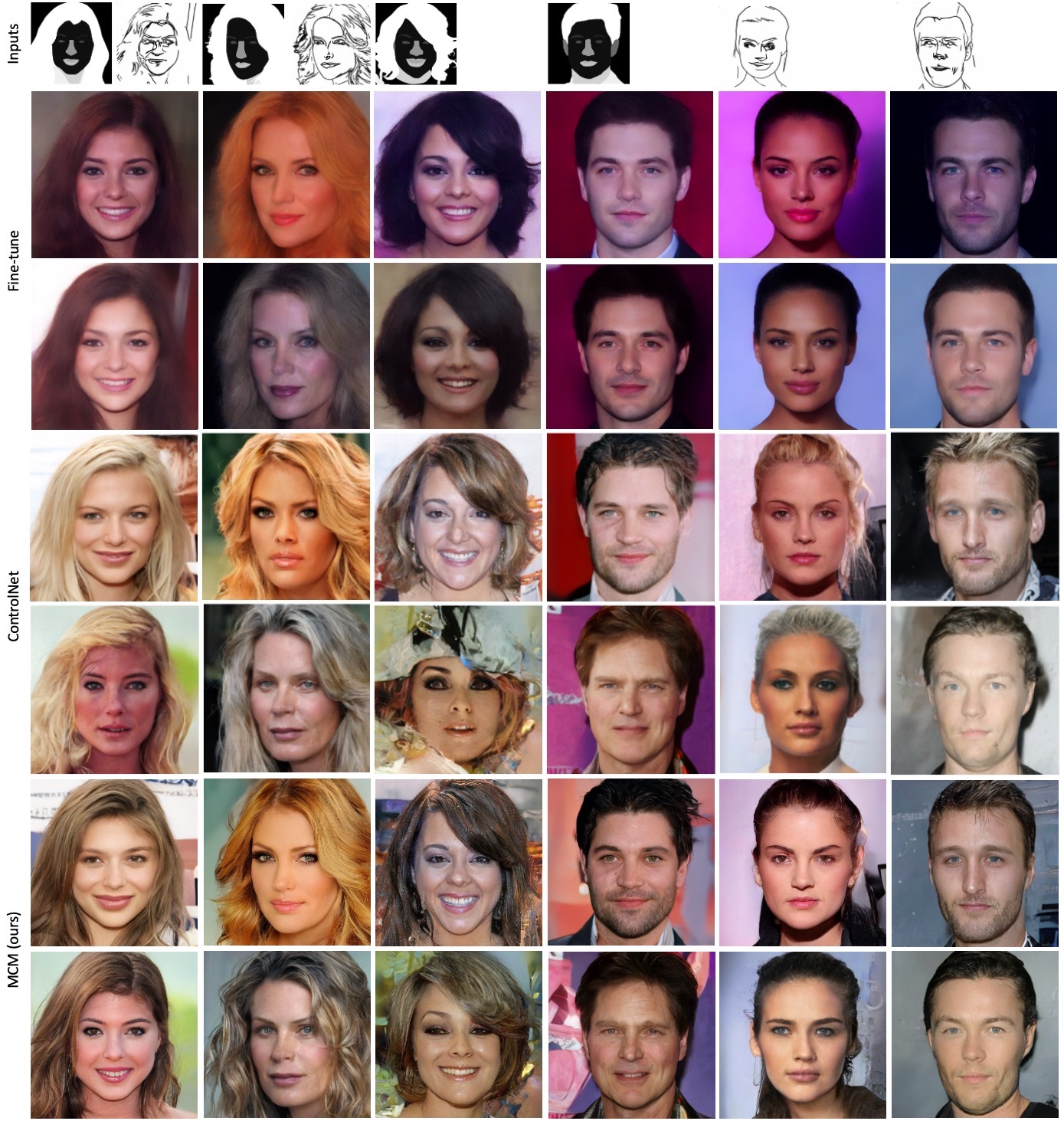}
  \caption{Qualitative comparison of fine-tuning and ControlNet against MCM for CelebA.}
  \label{fig:celeba-comparison}
\end{figure*}

\begin{figure*}
  \centering
  \includegraphics[width=0.61\textwidth]{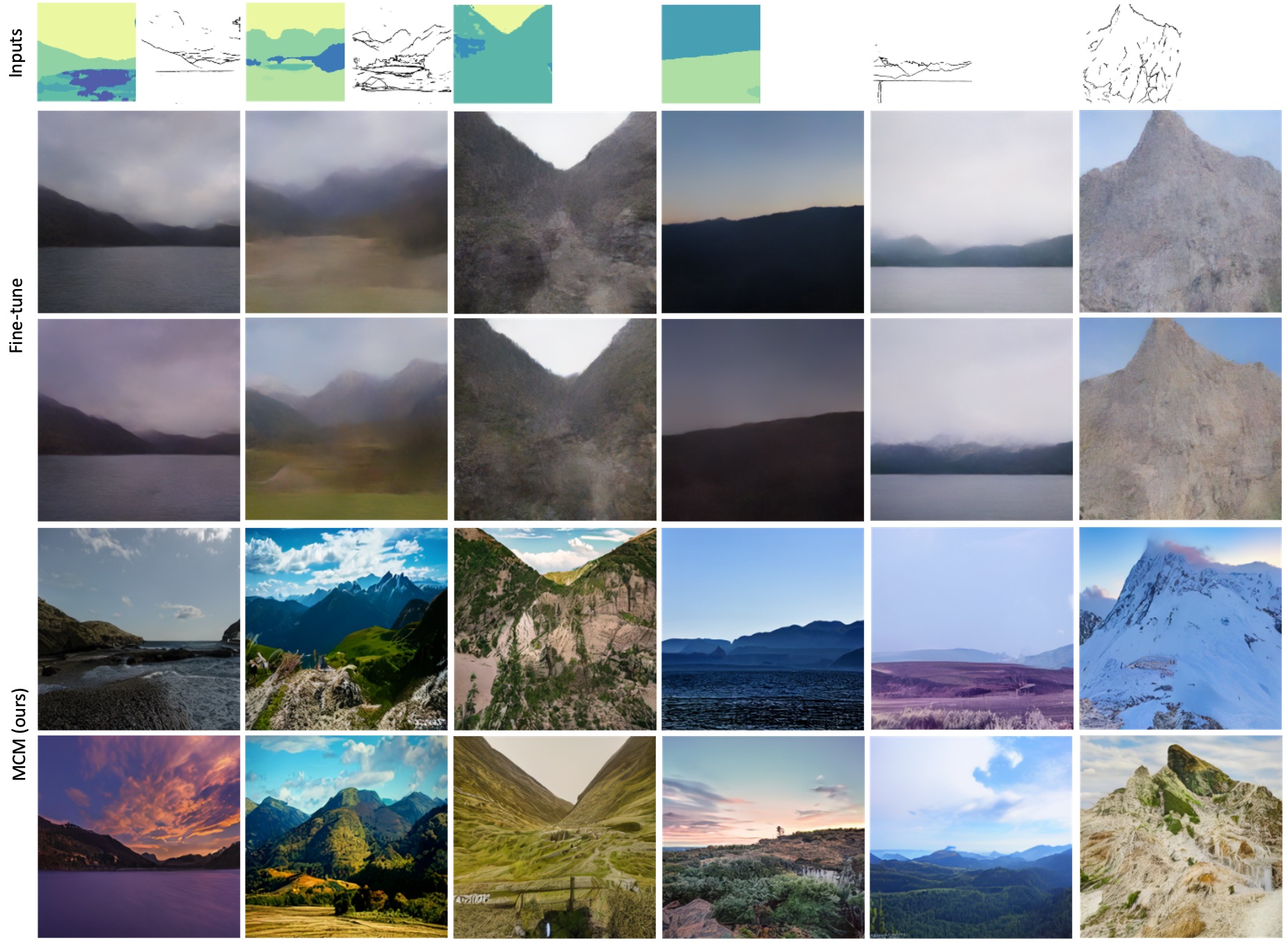}
  \caption{Qualitative comparison of the fine-tuning baseline against MCM for Mountains.}
  \label{fig:mountains-comparison}
\end{figure*}

\Cref{tab:seg-sketch-alignment} shows the alignment metrics for our method compared to all baselines. Compared to the base LDM we observe increased alignment between the inputs and generated images from MCM, fine-tuning, and ControlNet. Fine-tuning tends to produce blurrier images where the distinction between classes are unclear, which may account for the worse sketch alignment. Additionally, the identities of the faces generated by the fine-tuned CelebA model using the same inputs tend to be almost identical, relying mainly instead on illumination changes to produce ``diversity'' among the images. Thus, the greater alignment with fine-tuning comes at the expense of diversity (see \Cref{fig:celeba-comparison,fig:mountains-comparison}). ControlNet achieves similar alignment scores and quality as MCM.

We observe more difficulty with segmentation alignment on Mountains for MCM and fine-tuning. Unlike CelebA, where ground truth annotations are provided, the segmentation maps for Mountains are generated using an off-the-shelf network and span a larger number of classes (182 compared to CelebA's 19). Thus, both methods suffer from using poorer quality annotations as ground truth, but we believe that both would improve with better data. 

\begin{table}
  \begin{adjustbox}{width=\linewidth}
    \begin{tabular}{ccccccc}
      \toprule
      \multirow{2}{*}{Method}  &  \multicolumn{3}{c}{FID $\downarrow$}  &  \multicolumn{3}{c}{LPIPS $\uparrow$} \\
      &  Seg  &  Sketch  &  Seg + Sk  &  Seg  &  Sketch  &  Seg + Sk \\
      \midrule
      MSE  &  26.289  &  30.025  &  29.905  &  0.378  &  0.401  &  0.377 \\
      MSE + L1  &  18.085  &  21.065  &  18.842  &  0.458  &  0.479  &  0.461 \\
      MSE + L1 (full)  &  14.914  &  15.904  &  15.771  &  0.486  &  0.472  &  0.466 \\
      \bottomrule
    \end{tabular}
  \end{adjustbox}
  \caption{Ablation study using FID and LPIPS on CelebA.}
  \vspace{-2em}
  \label{tab:celeba-qualitative-ablation}
\end{table}

\begin{table}
  \begin{adjustbox}{width=\linewidth}
    \begin{tabular}{ccccccc}
      \toprule
      \multirow{2}{*}{Method}  &  \multicolumn{2}{c}{mIoU $\uparrow$}  &  \multicolumn{2}{c}{Accuracy $\uparrow$}  &  \multicolumn{2}{c}{Sketch dist. $\downarrow$} \\
      &  Seg  &  Seg + Sk  &  Seg  &  Seg + Sk  &  Sketch  &  Seg + Sk \\
      \midrule
      MSE  &  0.694  &  0.702  &  0.793  &  0.800  &  3.529  &  3.639 \\
      MSE + L1  &  0.631  &  0.627  &  0.744  &  0.741  &  4.323  &  4.485 \\
      MSE + L1 (full)  &  0.591  &  0.595  &  0.710  &  0.713  &  4.287  &  4.199 \\
      \bottomrule
    \end{tabular}
  \end{adjustbox}
  \caption{Ablation study using alignment metrics on CelebA.}
  \vspace{-2em}
  \label{tab:celeba-alignment-ablation}
\end{table}

\textbf{Ablation study.} We perform an ablation study using CelebA to evaluate the effect of the $L_1$ regularization term (\Cref{eq:l1}) and using limited training data. Results are shown in \Cref{tab:celeba-qualitative-ablation,tab:celeba-alignment-ablation}. We show that only using the MSE term for the training objective demonstrates similar behavior to the fine-tuning baseline--alignment improves while the overall quality and diversity of the images suffers. Thus, $L_1$ regularization of the modulation parameters helps balance the quality/consistency trade-off.

We compare MCM trained with a random subset of 5,000 examples against another trained with the full CelebA dataset. We train both modules with the same number of iterations so they observe the same number of training examples. Additional training data shows a similar pattern in the evaluation metrics to the addition of the $L_1$ term. The alignment metrics for MCM trained with the full dataset is likely to benefit from additional training time since there are more variations and more examples of the less common classes to learn. We also compare how ablating the amount of training data affects the quality of ControlNet and MCM in \Cref{sec:ablation-dataset-size}.


\section{Conclusion}
We introduce MCM, a novel method for multimodal image synthesis with diffusion models. Previous approaches to conditional image synthesis primarily rely on training from scratch or fine-tuning using large amounts of data and computational resources, which can be difficult or even infeasible. We avoid this by taking a pretrained diffusion model, freezing its weights, and only training a small module using a limited number of paired examples of new target modalities to modulate the sampling process. We evaluate our method using standard quality assessment metrics as well as alignment metrics to show that we are able to effectively incorporate user control while retaining high image quality.

\textbf{Limitations.} While our approach is able to efficiently apply multimodal control to pretrained diffusion models, MCM is currently limited to 2D modalities. We leave the incorporation of 1D modalities for future work, but show that MCM can be applied to text-conditioned models such as Stable Diffusion. Our approach can be more sensitive to the starting noise map $z_T$, and struggles with grounding semantics into class labels when the training data quality is poor. Additionally, MCM is limited to more structured domains.


\bibliographystyle{ACM-Reference-Format}
\bibliography{references}


\clearpage

\appendix

\section{Network architecture and training details} \label{sec:sup-reproducibility}

We use a time-conditional U-Net \cite{ronneberger2015u} for MCM and substitute the last convolutional layer with a split head, where one head outputs the multiplicative modulation parameter $\gamma$ and the other outputs the additive parameter $\nu$. The total number of parameters of MCM is $\sim$1\% of the unconditional LDMs and $\sim$0.4\% of SD. We use $\lambda_x = 1$ for unconditional LDMs and $\lambda_x = 10$ for SD. For all experiments, we use modality dropout rates $p_{seg} = p_{sketch} = 0.33$, and weighting term $\lambda_1 = \frac{1}{b \times h \times w \times c}$ in \Cref{eq:loss}, where $b$ is the batch size, and $(h, w, c)$ are the dimensions of the latent representations $z$. For SD, we randomly sample the latent vector from the KL-regularized autoencoder and modulate the output of the diffusion model without classifier-free guidance.

MCM takes in concatenated inputs $\{x_t, \epsilon_t, y_1, ..., y_n\}$ and timestep $t$, where $x_t, \epsilon_t \in \mathbb{R}^{h \times w \times c}$ and $y_i \in \mathbb{R}^{h \times w \times c_{y_i}}$ (for sketches and segmentation maps, $c_{y_1} = c_{y_2} = 1$). The last convolutional layer is replaced with a zero-initialized split head (one for outputting $\gamma$ and the other for $\nu$, where $\gamma, \nu \in \mathbb{R}^{h \times w \times c}$). For all experiments, MCM is trained with the hyperparameters described in \Cref{tab:sup-architecture} on NVIDIA A100s. We train 3 MCMs--one MCM per diffusion model/dataset combination (unconditional CelebA LDM, unconditional Mountains LDM, text-conditioned Mountains SD).

\begin{table}[]
  \centering
  \begin{tabular}{cc}
    \toprule
    \# parameters  &  3.9M \\
    Channels  &  32 \\
    Channel multiplier  &  1,1,2,4 \\
    \# residual blocks  &  2 \\
    Attention resolutions  &  16 \\
    Batch size  &  64 \\
    Epochs  &  10k \\
    Learning rate  &  1e-5 \\
    \bottomrule
  \end{tabular}
  \caption{MCM hyperparameters for all experiments.}
  \vspace{-1em}
  \label{tab:sup-architecture}
\end{table}

\section{MCM with Stable Diffusion} \label{sec:sup-sd}

\begin{figure*}
  \centering
  \includegraphics[height=0.9\textheight]{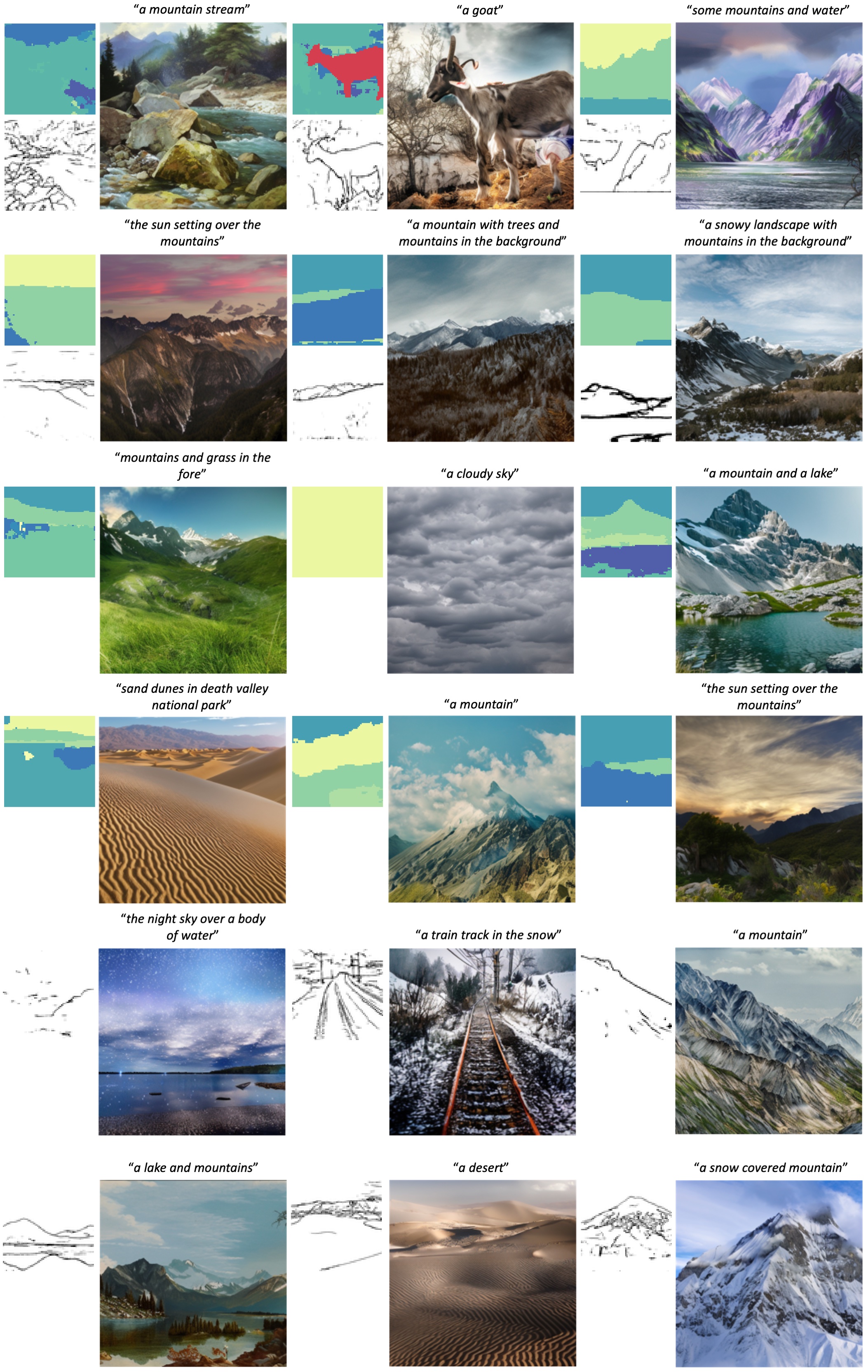}
  \caption{Examples generated by MCM with SD for Mountains.}
  \label{fig:sup-mountains-sd-ddim}
\end{figure*}

\begin{figure*}
  \centering
  \includegraphics[height=0.9\textheight]{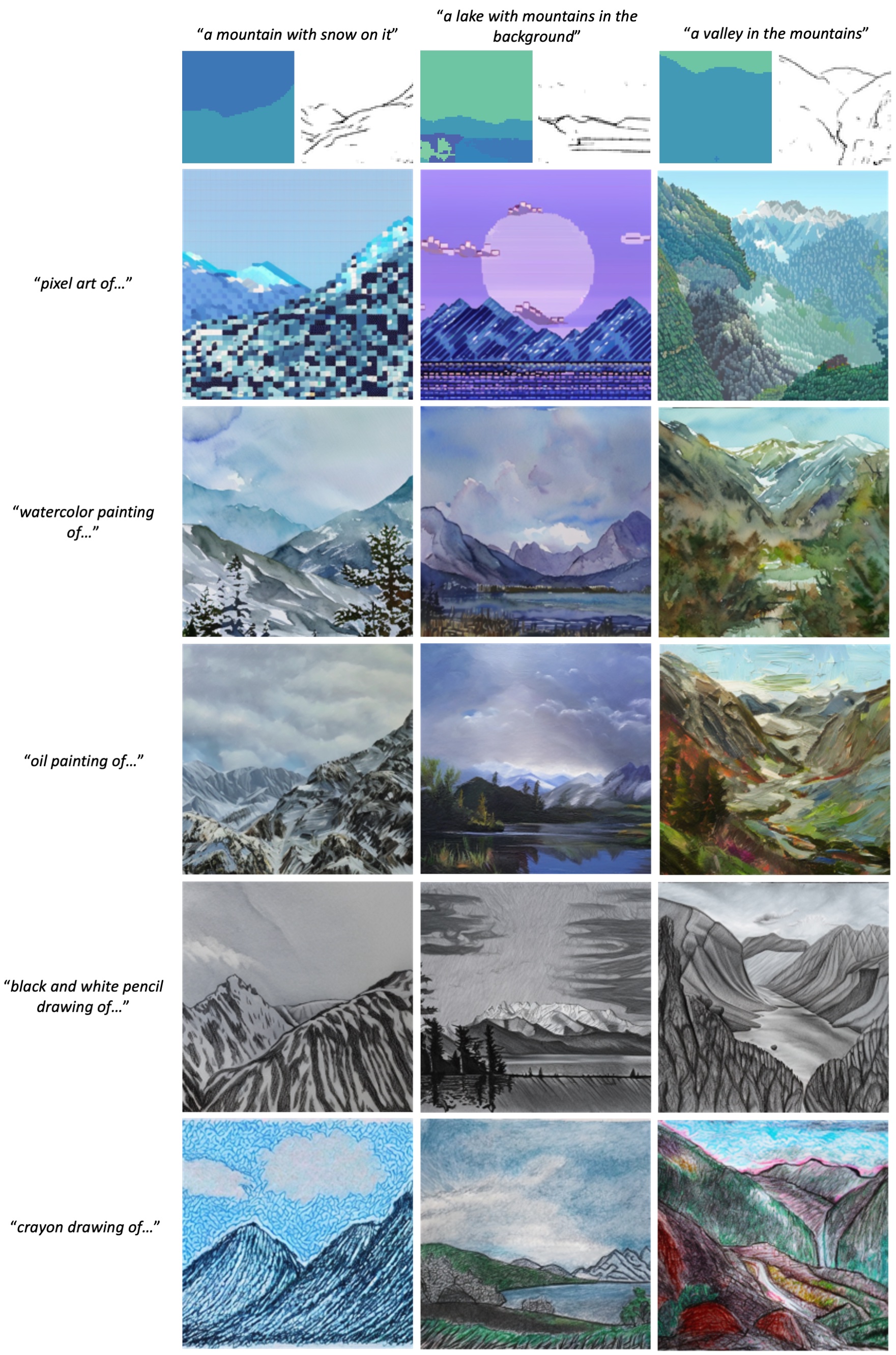}
  \caption{Varying the artistic style using segmentation map and sketch inputs to MCM with SD.}
  \label{fig:sup-sd-styles-segmap-sketch}
\end{figure*}

\begin{figure*}
  \centering
  \includegraphics[height=0.9\textheight]{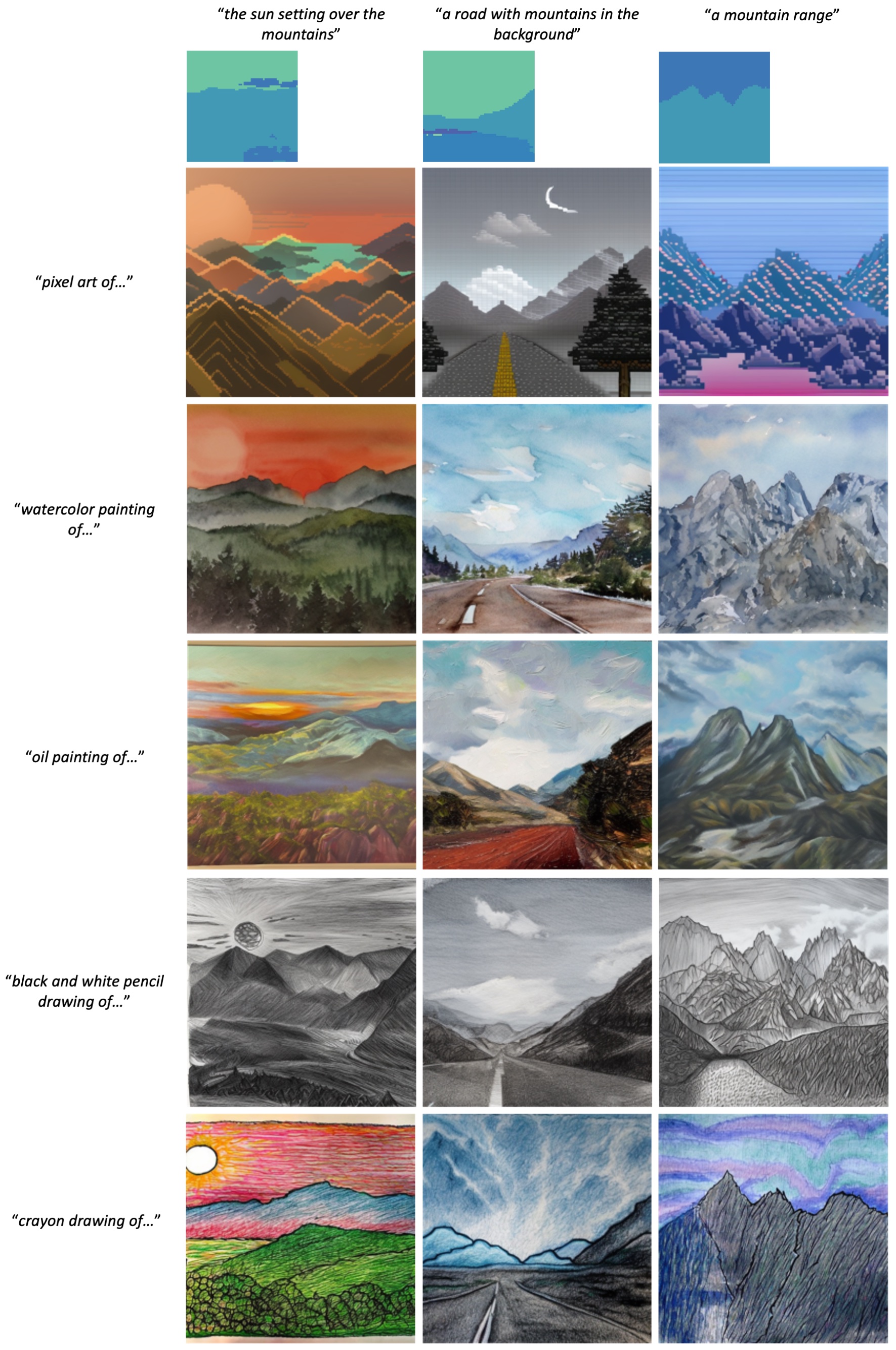}
  \caption{Varying the artistic style using segmentation map inputs to MCM with SD.}
  \label{fig:sup-sd-styles-segmap}
\end{figure*}

\begin{figure*}
  \centering
  \includegraphics[height=0.9\textheight]{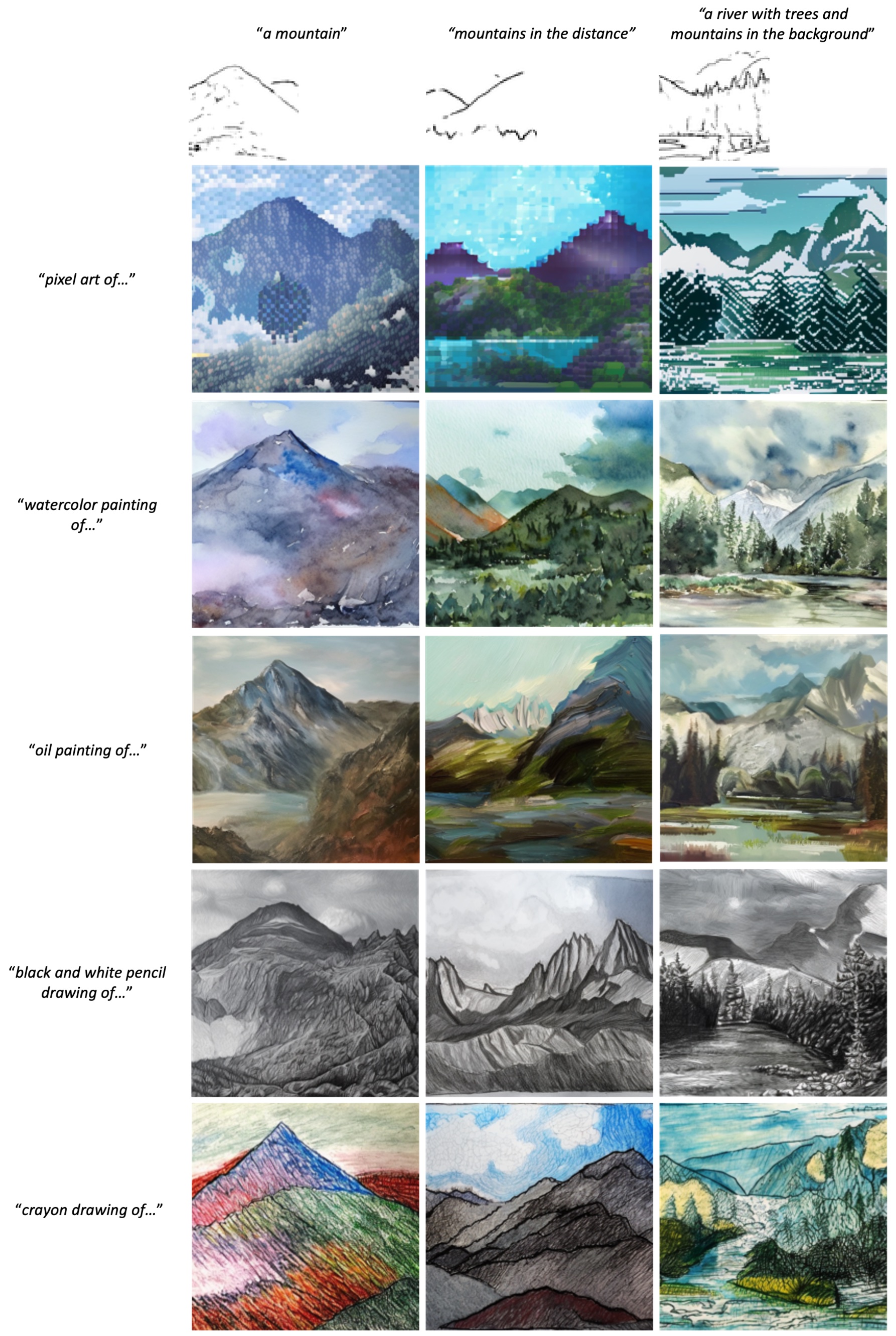}
  \caption{Varying the artistic style using sketch inputs to MCM with SD.}
  \label{fig:sup-sd-styles-sketch}
\end{figure*}

We provide additional examples of applying MCM to Stable Diffusion v2.1 using DDIM with 200 steps and an unconditional guidance scale of 5.0 in \Cref{fig:sup-mountains-sd-ddim}. We also experiment with varying the artistic styles through keywords in the text input to SD in \Cref{fig:sup-sd-styles-segmap-sketch,fig:sup-sd-styles-segmap,fig:sup-sd-styles-sketch}. 

\section{Sampling methods} \label{sec:sup-sampling}

\begin{figure*}
  \centering
  \includegraphics[height=0.9\textheight]{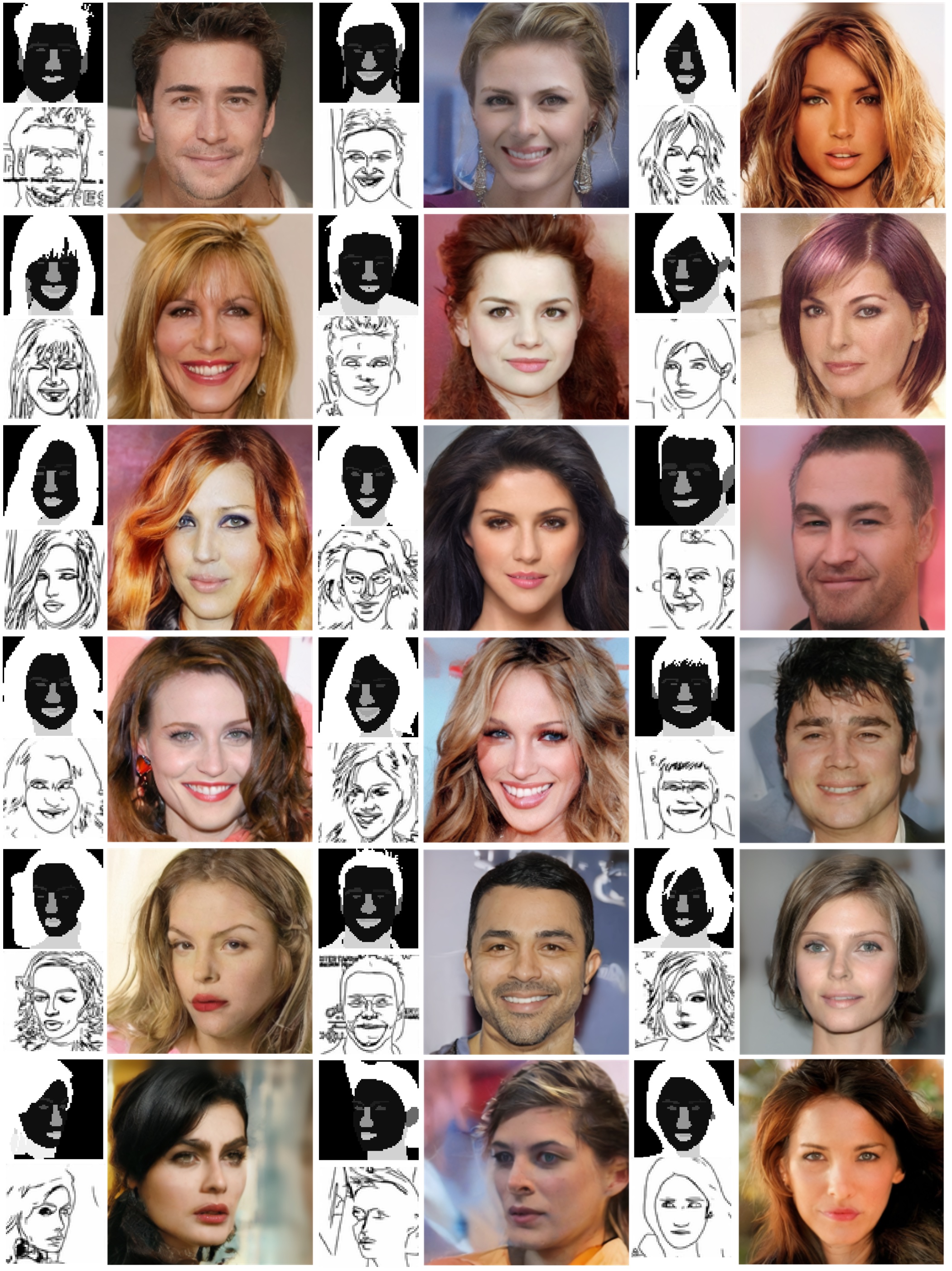}
  \caption{Examples generated by MCM using DDIM sampling with segmentation maps and sketches on CelebA.}
  \label{fig:sup-celeba-segmap-sketch-ddim}
\end{figure*}

\begin{figure*}
  \centering
  \includegraphics[height=0.9\textheight]{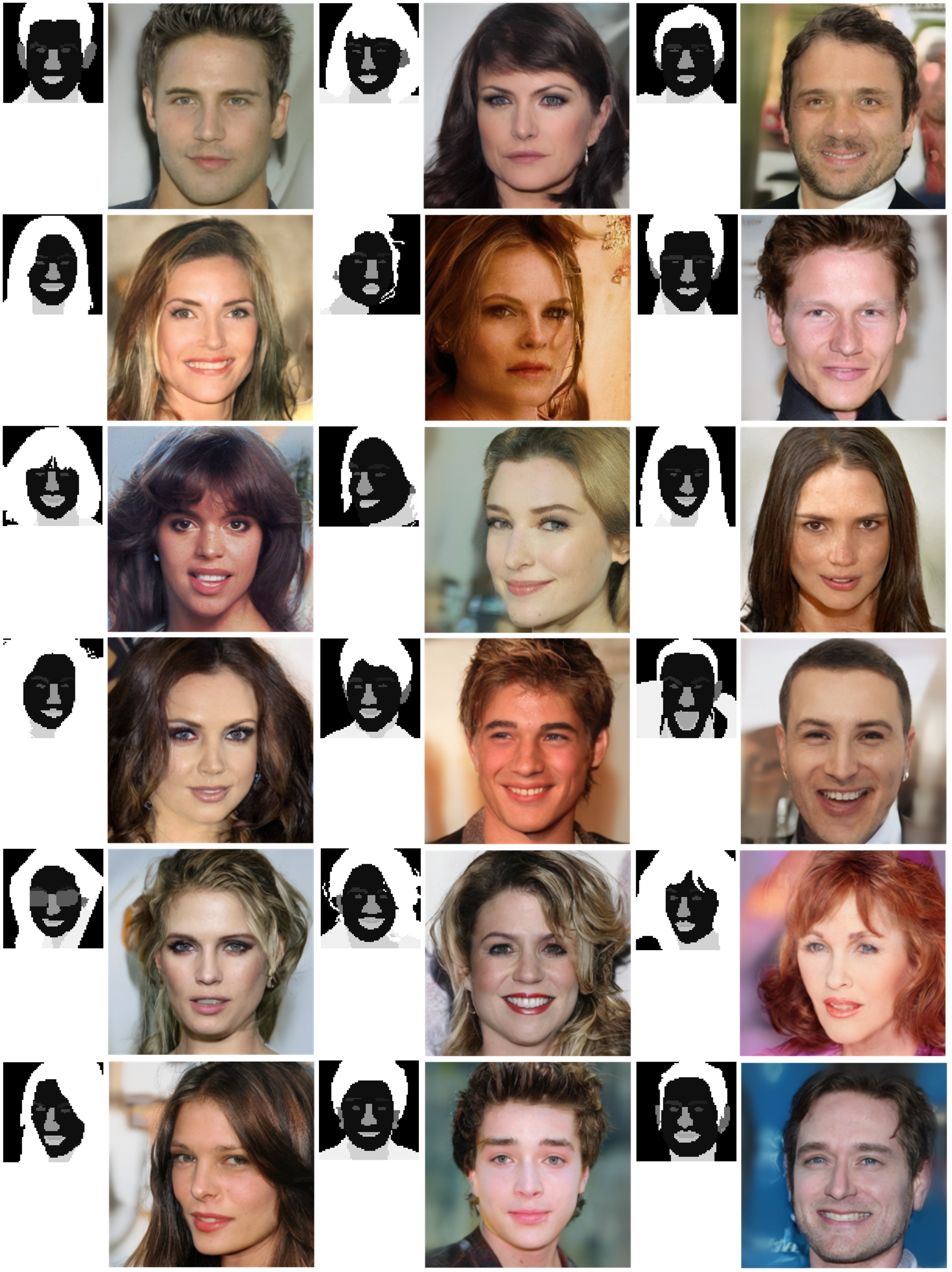}
  \caption{Examples generated by MCM using DDIM sampling with segmentation maps on CelebA.}
  \label{fig:sup-celeba-segmap-ddim}
\end{figure*}

\begin{figure*}
  \centering
  \includegraphics[height=0.9\textheight]{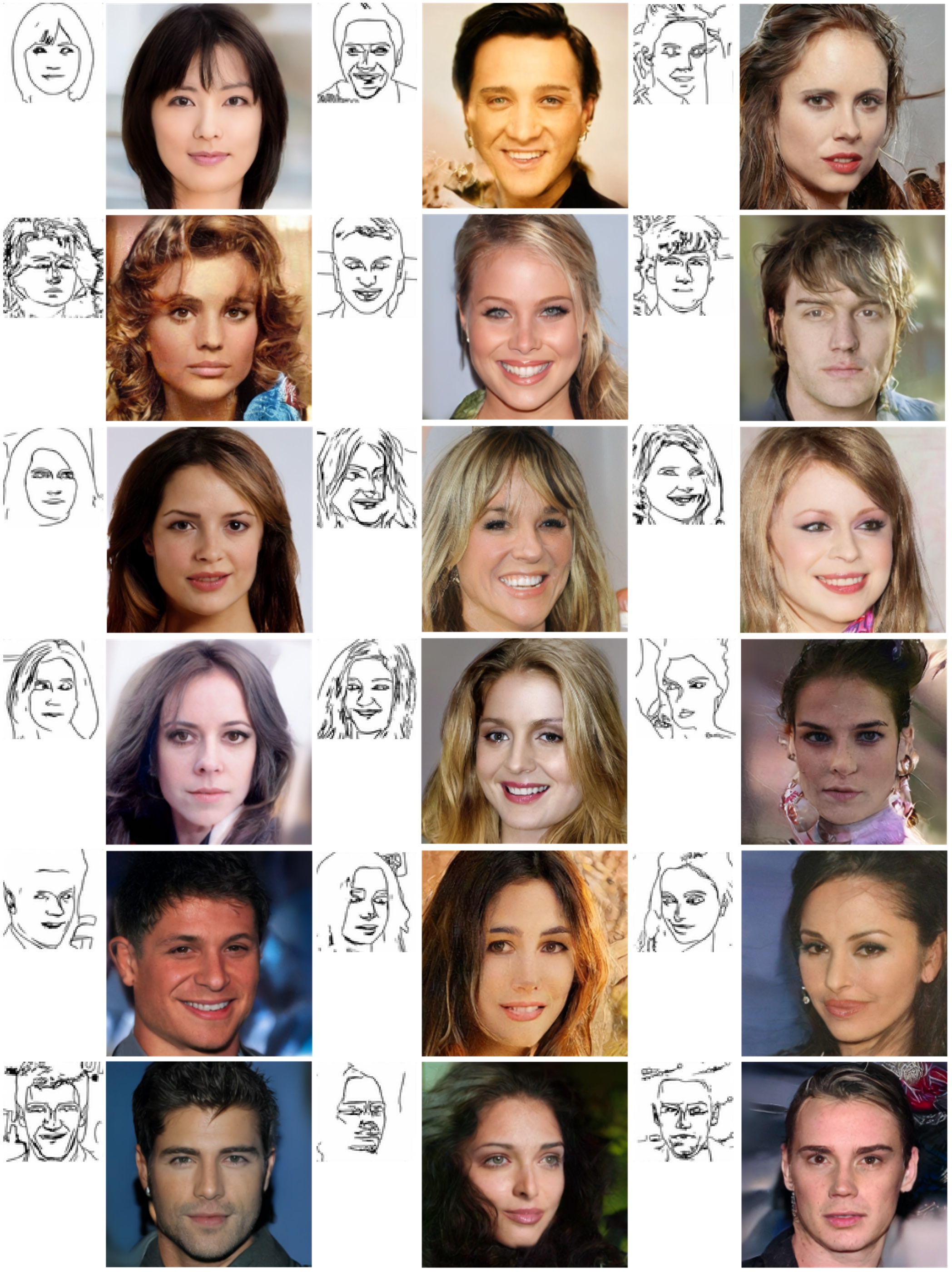}
  \caption{Examples generated by MCM using DDIM sampling with sketches on CelebA.}
  \label{fig:sup-celeba-sketch-ddim}
\end{figure*}

\begin{figure*}
  \centering
  \includegraphics[height=0.9\textheight]{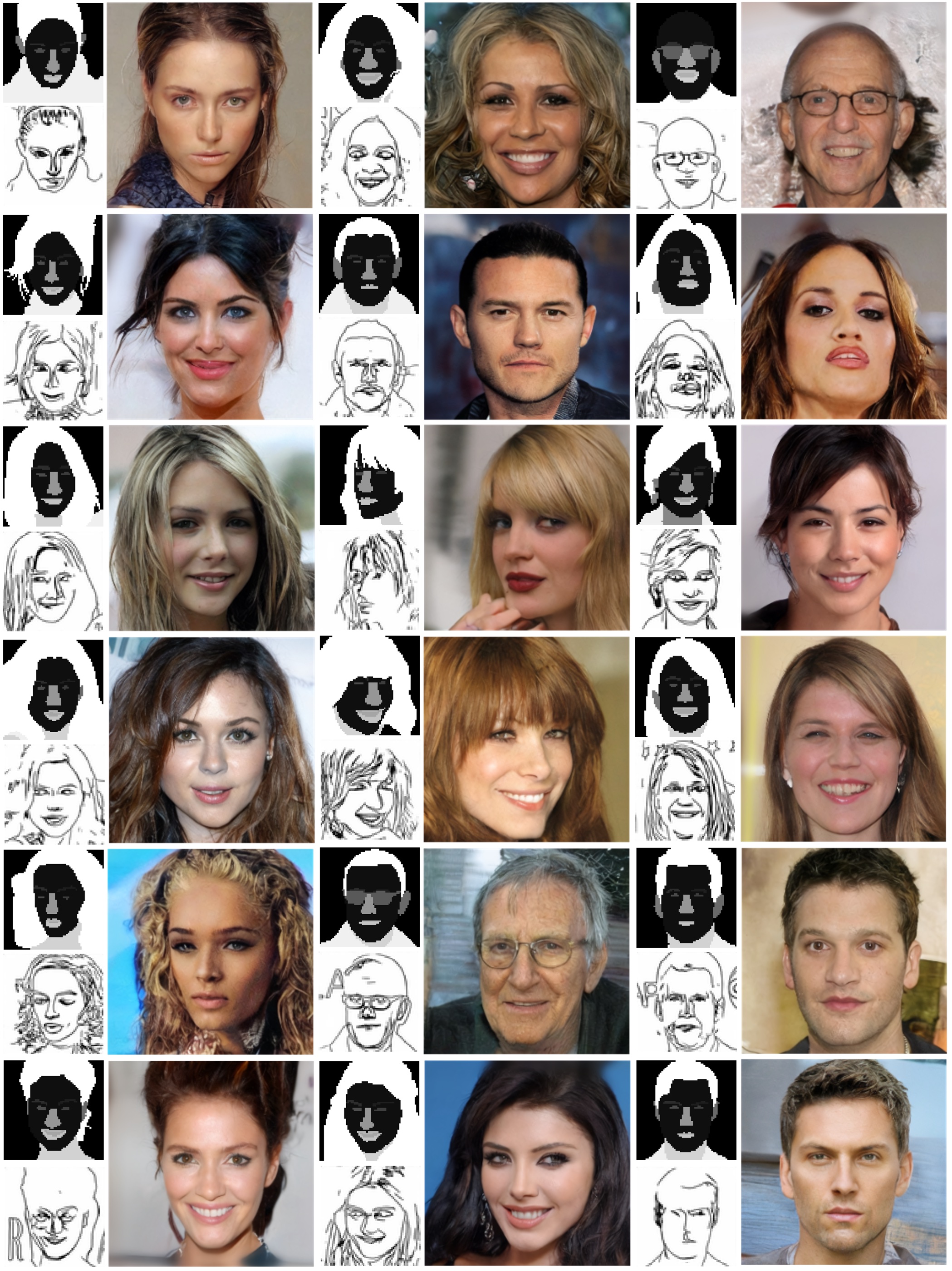}
  \caption{Examples generated by MCM using DDPM sampling with segmentation maps and sketches on CelebA.}
  \label{fig:sup-celeba-segmap-sketch-ddpm}
\end{figure*}

\begin{figure*}
  \centering
  \includegraphics[height=0.9\textheight]{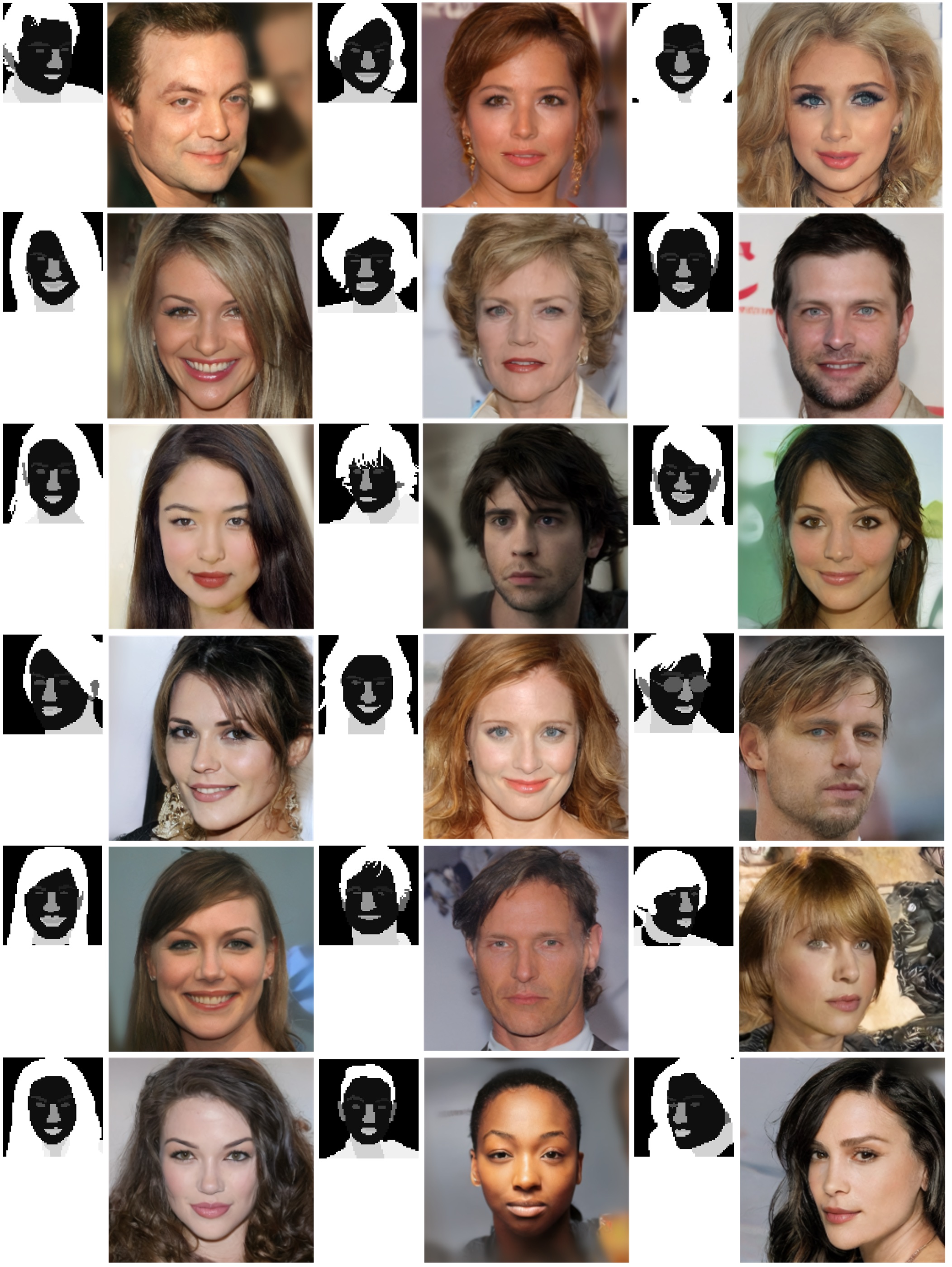}
  \caption{Examples generated by MCM using DDPM sampling with segmentation maps on CelebA.}
  \label{fig:sup-celeba-segmap-ddpm}
\end{figure*}

\begin{figure*}
  \centering
  \includegraphics[height=0.9\textheight]{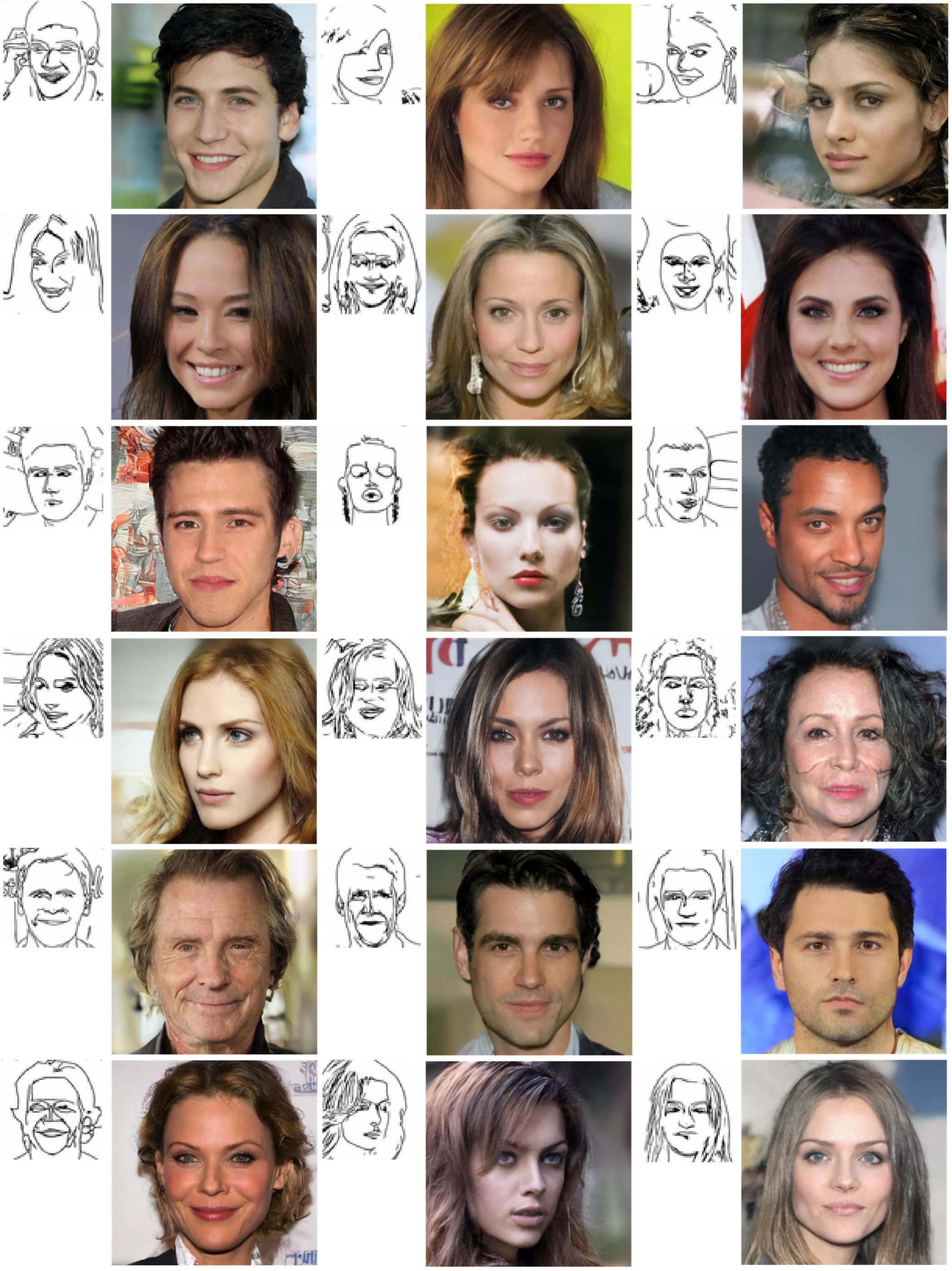}
  \caption{Examples generated by MCM using DDPM sampling with sketches on CelebA.}
  \label{fig:sup-celeba-sketch-ddpm}
\end{figure*}

\begin{figure*}
  \centering
  \includegraphics[height=0.9\textheight]{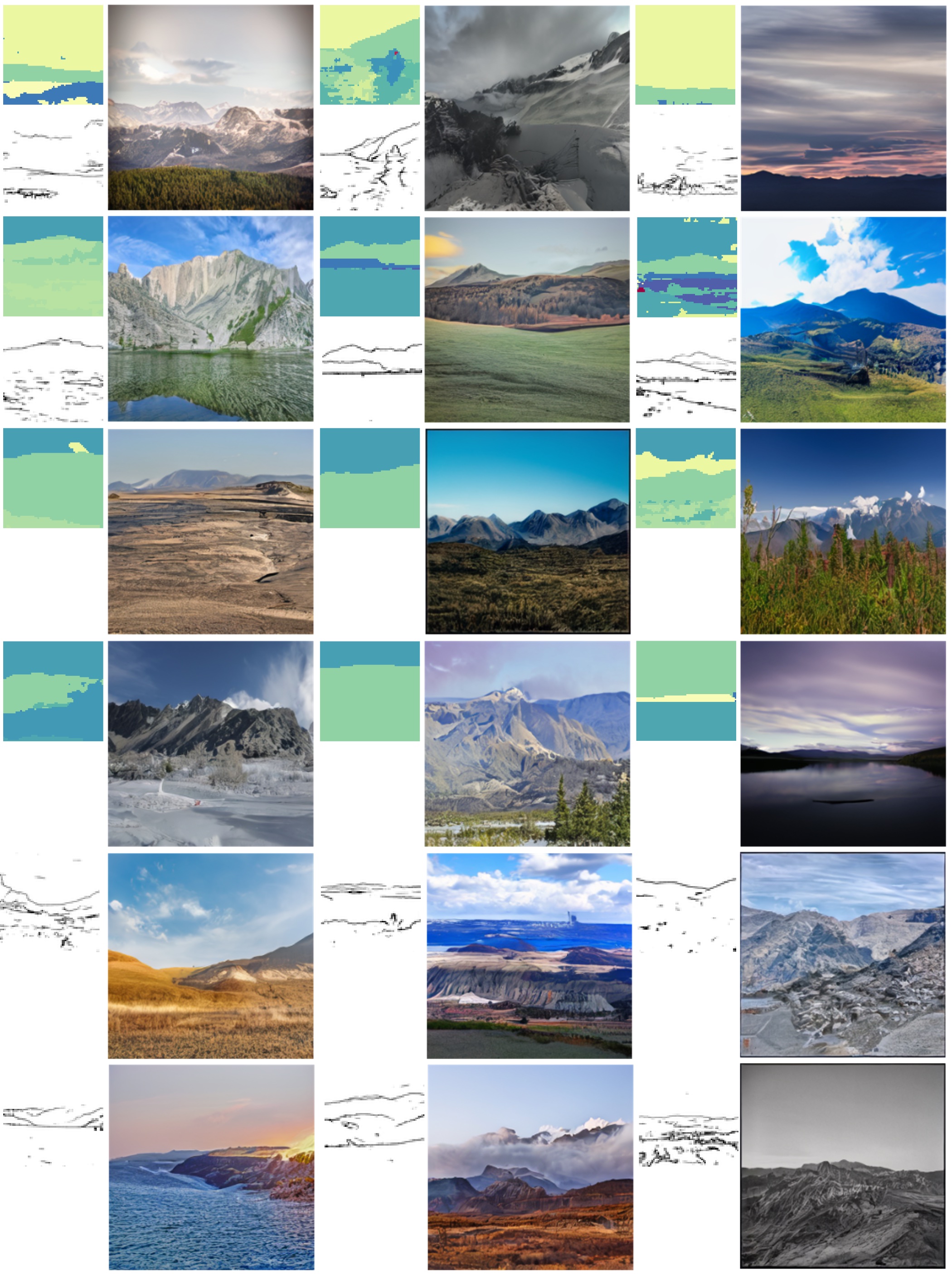}
  \caption{Examples generated by MCM using DDIM sampling on Mountains.}
  \label{fig:sup-mountains-ddim}
\end{figure*}

\begin{figure*}
  \centering
  \includegraphics[height=0.9\textheight]{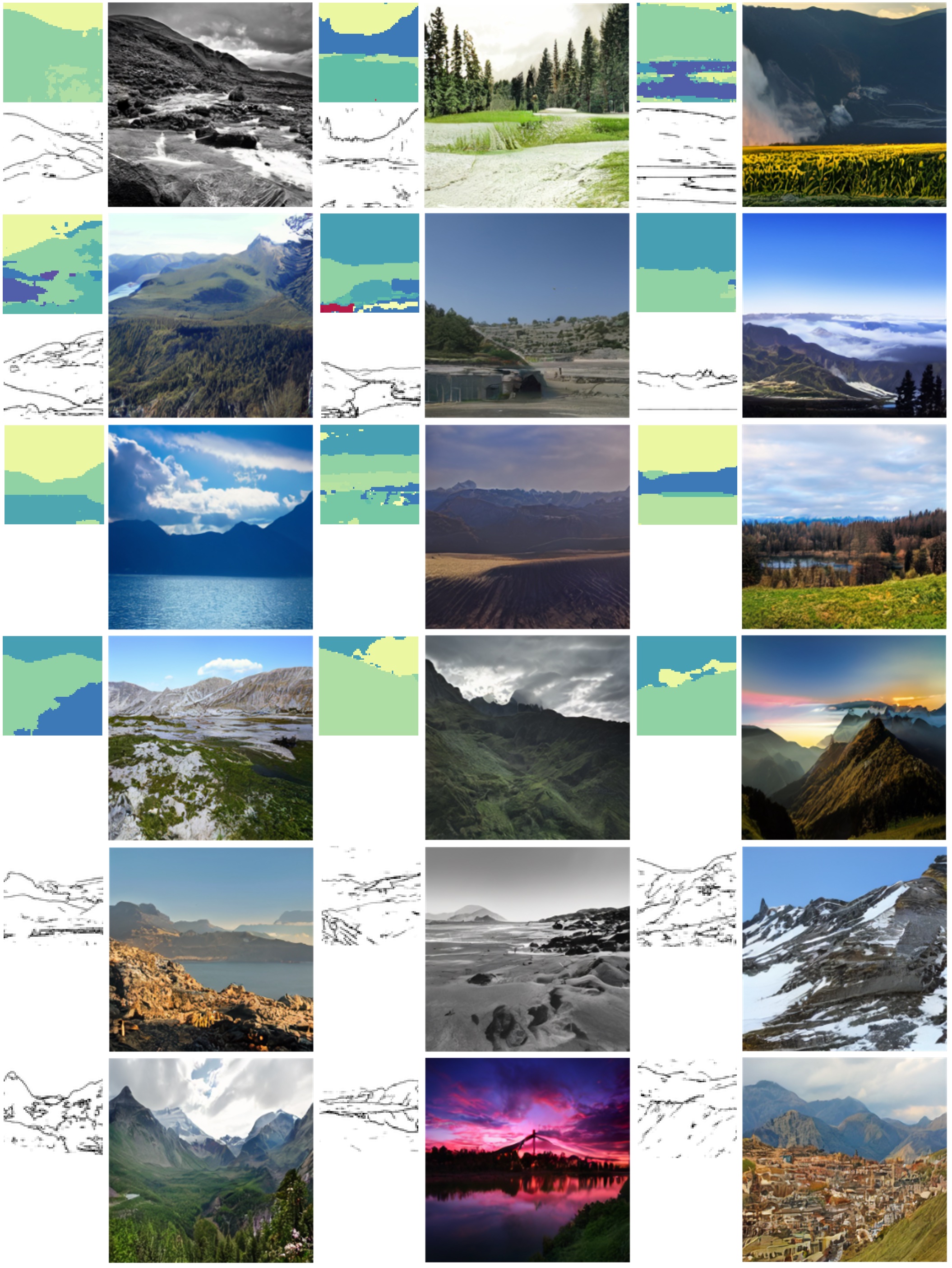}
  \caption{Examples generated by MCM using DDPM sampling on Mountains.}
  \label{fig:sup-mountains-ddpm}
\end{figure*}

Since MCM is trained to predict modulation parameters for each timestep independently, we are flexible in the choice of sampling technique. We provide examples using DDPM \cite{ho2020denoising} with all 1000 steps (\Cref{fig:sup-celeba-segmap-sketch-ddpm,fig:sup-celeba-segmap-ddpm,fig:sup-celeba-sketch-ddpm} for CelebA and \Cref{fig:sup-mountains-ddpm} for Mountains) as well as additional results using DDIM \cite{song2020denoising} with 200 uniformly sampled steps (\Cref{fig:sup-celeba-segmap-sketch-ddim,fig:sup-celeba-segmap-ddim,fig:sup-celeba-sketch-ddim} for CelebA and \Cref{fig:sup-mountains-ddim} for Mountains).

We omit results using PLMS \cite{liu2022pseudo} because of the high similarity to the samples produced with DDIM when using the same input noise $z_T$.

\section{Effect of dataset size on MCM and ControlNet} \label{sec:ablation-dataset-size}

In \Cref{tab:celeba-num-qualitative-ablation,tab:celeba-num-alignment-ablation}, we further reduce the number of training examples and compare against ControlNet \cite{zhang2023adding}. ControlNet is concurrent work aims to add new conditioning modalities to pretrained diffusion models by training a copy of the diffusion model's weights. The ``trainable copy'' is used to modulate the features of the original ``locked copy'', and thus can be seen as a variant of our approach with direct access to the diffusion model and better initialization. We modify ControlNet similarly to MCM in order accommodate multimodal synthesis (e.g., concatenating modalities, using modality dropout). We evaluate the best overall checkpoints for each ControlNet to MCM, which were all trained for the same number of epochs, and find that the two methods perform comparably even though ControlNet has significantly more trainable parameters. We observe that reducing the number of training examples for ControlNet leads to both poorer quality and alignment. Meanwhile, reducing the training examples for MCM produces more photorealistic and diverse images, but the images have poorer alignment to the input conditions.

\begin{table}
  \begin{adjustbox}{width=\linewidth}
    \begin{tabular}{cccccccc}
      \toprule
      \multirow{2}{*}{Method}  &  \multirow{2}{*}{\# train ex.}  &  \multicolumn{3}{c}{FID $\downarrow$}  &  \multicolumn{3}{c}{LPIPS $\uparrow$} \\
      &  &  Seg  &  Sketch  &  Seg + Sk  &  Seg  &  Sketch  &  Seg + Sk \\
      \midrule
      \multirow{4}{*}{ControlNet}  &  500  &  21.278  &  22.576  &  21.076  &  0.500  &  0.545  &  0.498 \\
      &  1000  &  21.487  &  21.429  &  21.305  &  0.502  &  0.546  &  0.498 \\
      &  2500  &  18.979  &  18.739  &  18.759  &  0.498  &  0.546  &  0.493 \\
      &  5000  &  19.317  &  17.638  &  18.875  &  0.496  &  0.546  &  0.492 \\
      \midrule
      \multirow{4}{*}{MCM (Ours)}  &  500  &  17.549  &  17.595  &  16.781  &  0.514  &  0.527  &  0.511 \\
      &  1000  &  17.373  &  19.038  &  17.797  &  0.499  &  0.518  &  0.498 \\
      &  2500  &  17.456  &  19.464  &  18.368  &  0.498  &  0.500  &  0.484 \\
      &  5000  &  18.085  &  21.065  &  18.842  &  0.458  &  0.479  &  0.461 \\
      \bottomrule
    \end{tabular}
  \end{adjustbox}
  \caption{Ablation study over training set size using FID and LPIPS on CelebA.}
  \vspace{-2em}
  \label{tab:celeba-num-qualitative-ablation}
\end{table}

\begin{table}
  \begin{adjustbox}{width=\linewidth}
    \begin{tabular}{cccccccc}
      \toprule
      \multirow{2}{*}{Method}  &  \multirow{2}{*}{\# train ex.}  &  \multicolumn{2}{c}{mIoU $\uparrow$}  &  \multicolumn{2}{c}{Accuracy $\uparrow$}  &  \multicolumn{2}{c}{Sketch dist. $\downarrow$} \\
      &  &  Seg  &  Seg + Sk  &  Seg  &  Seg + Sk  &  Sketch  &  Seg + Sk \\
      \midrule
      \multirow{4}{*}{ControlNet}  &  500  &  0.553  &  0.557  &  0.673  &  0.677  &  7.811  &  6.526 \\
      &  1000  &  0.557  &  0.564  &  0.675  &  0.680  &  7.403  &  6.304 \\
      &  2500  &  0.584  &  0.590  &  0.697  &  0.701  &  7.596  &  6.277 \\
      &  5000  &  0.616  &  0.623  &  0.723  &  0.728  &  7.592  &  6.072 \\
      \midrule
      \multirow{4}{*}{MCM (Ours)}  &  500  &  0.533  &  0.537  &  0.660  &  0.664  &  5.361  &  5.267 \\
      &  1000  &  0.559  &  0.559  &  0.683  &  0.682  &  5.148  &  5.023 \\
      &  2500  &  0.608  &  0.583  &  0.726  &  0.701  &  4.773  &  4.612 \\
      &  5000  &  0.631  &  0.627  &  0.744  &  0.741  &  4.323  &  4.485 \\
      \bottomrule
    \end{tabular}
  \end{adjustbox}
  \caption{Ablation study over training set size using alignment metrics on CelebA.}
  \vspace{-2em}
  \label{tab:celeba-num-alignment-ablation}
\end{table}


\section{MCM with pixel-based diffusion models} \label{sec:sup-pixel}

\begin{figure*}
  \centering
  \includegraphics[height=0.9\textheight]{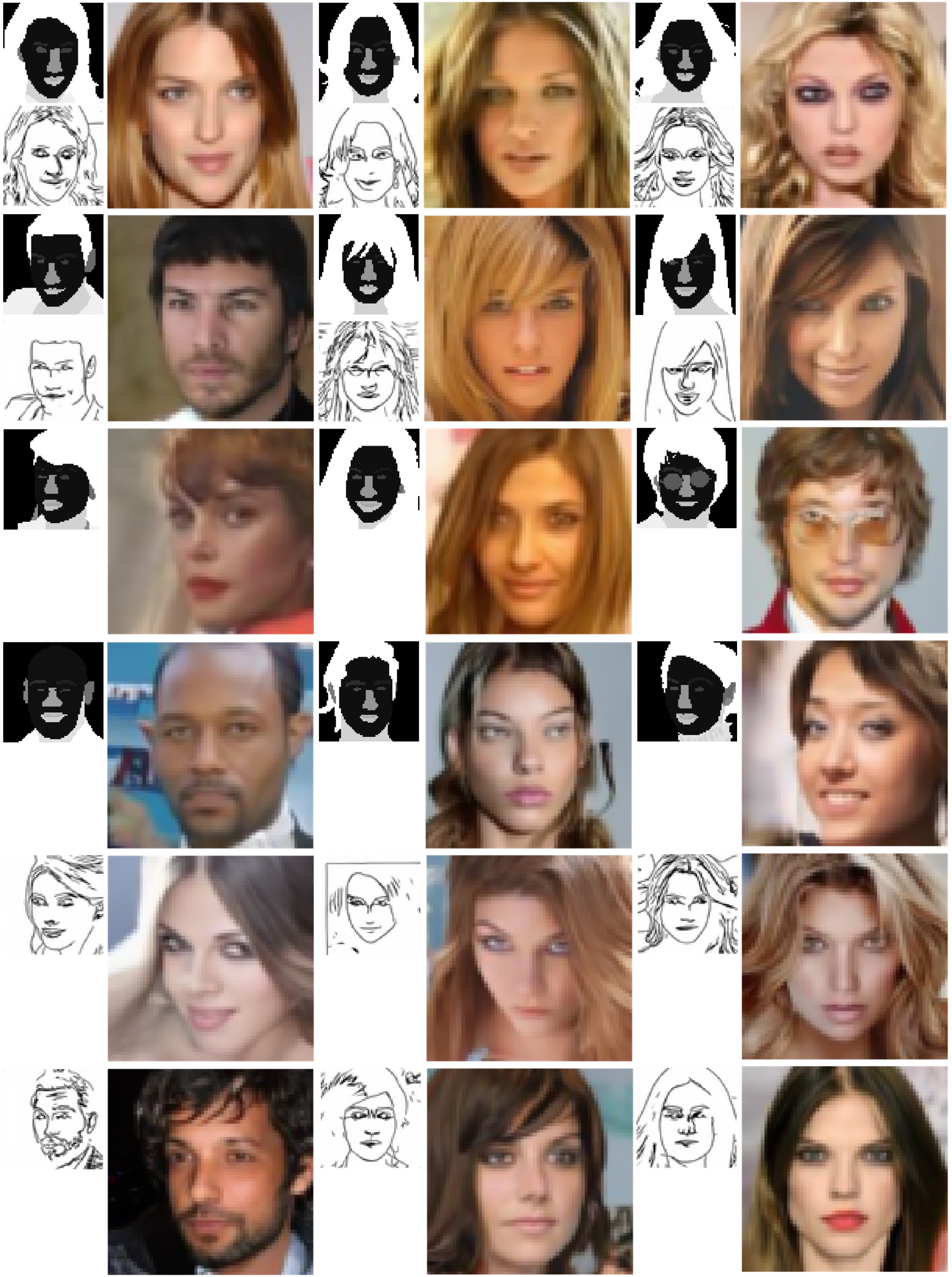}
  \caption{Examples generated by applying MCM to a pixel-based diffusion model.}
  \label{fig:sup-celeba-pixel}
\end{figure*}

We apply MCM to a pixel-based diffusion model in \Cref{fig:sup-celeba-pixel}. We use a public checkpoint for an unconditional model trained on CelebA at $64\times64$ resolution\footnote{\url{https://github.com/ermongroup/ddim}}. We use the same architecture and setup as described in \Cref{sec:experiments} with one small modification: before applying \Cref{eq:mse-loss} to the predicted denoised image $x_0'$, we use \textit{static thresholding} on $x_0'$ by clipping the values to $[-1,1]$.

\end{document}